\documentclass[10pt,journal,compsoc]{IEEEtran}
\usepackage{graphicx}
\usepackage{subcaption}
\usepackage[table]{xcolor}% http://ctan.org/pkg/xcolor
\usepackage{url}
\usepackage{algorithm}
\usepackage[noend]{algpseudocode}
\usepackage{xspace}
\usepackage{booktabs}
\usepackage{multirow}
\usepackage{siunitx}

\usepackage{amsmath,amssymb}
\usepackage{mathrsfs}
\usepackage{mathtools}
\usepackage{ragged2e}
\usepackage{wrapfig}
\newcommand{\hlcyan}[1]{{#1}}

\usepackage[pagebackref=true,breaklinks=true,letterpaper=true,colorlinks,bookmarks=false]{hyperref}

\ifCLASSOPTIONcompsoc
  \usepackage[nocompress]{cite}
\else
  \usepackage{cite}
\fi
\ifCLASSINFOpdf
\else
\fi
\begin{document}
\title{On Symbiosis of Attribute Prediction and Semantic Segmentation}

\author{Mahdi~M.~Kalayeh,~\IEEEmembership{Member,~IEEE,}
        and~Mubarak~Shah,~\IEEEmembership{Fellow,~IEEE}% <-this % stops a space
\IEEEcompsocitemizethanks{\IEEEcompsocthanksitem At the time of the initial submission, M. M. Kalayeh and M. Shah were with the Center for Research in Computer Vision, University of Central Florida, Orlando,
FL, 32816.\protect\\
E-mails: mahdi@eecs.ucf.edu, shah@crcv.ucf.edu\protect\\
This work updates and extends our previous work \cite{kalayeh2017improving}.
}% <-this % stops an unwanted space
\thanks{}}

\IEEEtitleabstractindextext{%
\begin{abstract}
\justifying
Attributes are semantically meaningful characteristics whose applicability widely crosses category boundaries. They are particularly important in describing and recognizing concepts for which no explicit training example is given, \textit{e.g., zero-shot learning}. Additionally, since attributes are human describable, they can be used for efficient human-computer interaction. In this paper, we propose to employ semantic segmentation to improve person-related attribute prediction. The core idea lies in the fact that many attributes describe local properties. In other words, the probability of an attribute to appear in an image is far from being uniform in the spatial domain. We build our attribute prediction model jointly with a deep semantic segmentation network. This harnesses the localization cues learned by the semantic segmentation to guide the attention of the attribute prediction to the regions where different attributes naturally show up. As a result of this approach, in addition to prediction, we are able to localize the attributes despite merely having access to image-level labels (weak supervision) during training. We first propose semantic segmentation-based pooling and gating, respectively denoted as SSP and SSG. In the former, the estimated segmentation masks are used to pool the final activations of the attribute prediction network, from multiple semantically homogeneous regions. This is in contrast to global average pooling which is agnostic with respect to where in the spatial domain activations occur. In SSG, the same idea is applied to the intermediate layers of the network. Specifically, we create multiple copies of the internal activations. In each copy, only values that fall within a certain semantic region are preserved while outside of that, activations are suppressed. This mechanism allows us to prevent pooling operation from blending activations that are associated with semantically different regions. SSP and SSG, while effective, impose heavy memory utilization since each channel of the activations is pooled/gated with \textit{all} the semantic segmentation masks. To circumvent this, we propose Symbiotic Augmentation (SA), where we \textit{learn} only one mask per activation channel. SA allows the model to either pick one, or combine (weighted superposition) multiple semantic maps, in order to generate the proper mask for each channel. SA simultaneously applies the same mechanism to the reverse problem by leveraging output logits of attribute prediction to guide the semantic segmentation task. We evaluate our proposed methods for facial attributes on CelebA  and LFWA datasets, while benchmarking WIDER Attribute and Berkeley Attributes of People for whole body attributes. Our proposed methods achieve superior results compared to the previous works. Furthermore, we show that in the reverse problem, semantic face parsing significantly improves when its associated task is jointly learned, through our proposed Symbiotic Augmentation (SA), with facial attribute prediction. We confirm that when few training instances are available, indeed image-level facial attribute labels can serve as an effective source of weak supervision to improve semantic face parsing. That reaffirms the need to jointly model these two interconnected tasks.
\end{abstract}

\begin{IEEEkeywords}
Attribute Prediction, Semantic Segmentation, Semantic Gating, Facial Attributes, Person Attributes 
\end{IEEEkeywords}}

\maketitle

\IEEEdisplaynontitleabstractindextext

% \fi
\IEEEpeerreviewmaketitle

\IEEEraisesectionheading{\section{Introduction}\label{sec:introduction}}

\IEEEPARstart{N}{owadays}, state-of-the-art computer vision techniques allow us to teach machines different classes of objects, actions, scenes, and even fine-grained categories. However, to learn a certain notion, we usually need positive and negative examples from the concept of interest. This creates a set of challenges as the instances of different concepts are not equally easy to collect. Also, the number of learnable concepts is linearly capped by the cardinality of the training data. Therefore, being able to robustly learn a set of \textit{sharable concepts} that go beyond rigid category boundaries is of tremendous importance. Visual attributes are one particular type of these \textit{sharable concepts}. They are human describable and machine detectable. We can use attributes to describe a variety of objects, scenes, actions, and events. For example, we associate a person who is lying on a beach with the attribute \textit{relaxed} or a cat that is chasing after a wool ball with the attribute \textit{playing}.

\begin{figure}
	\centering
	\includegraphics[width=0.48\textwidth]{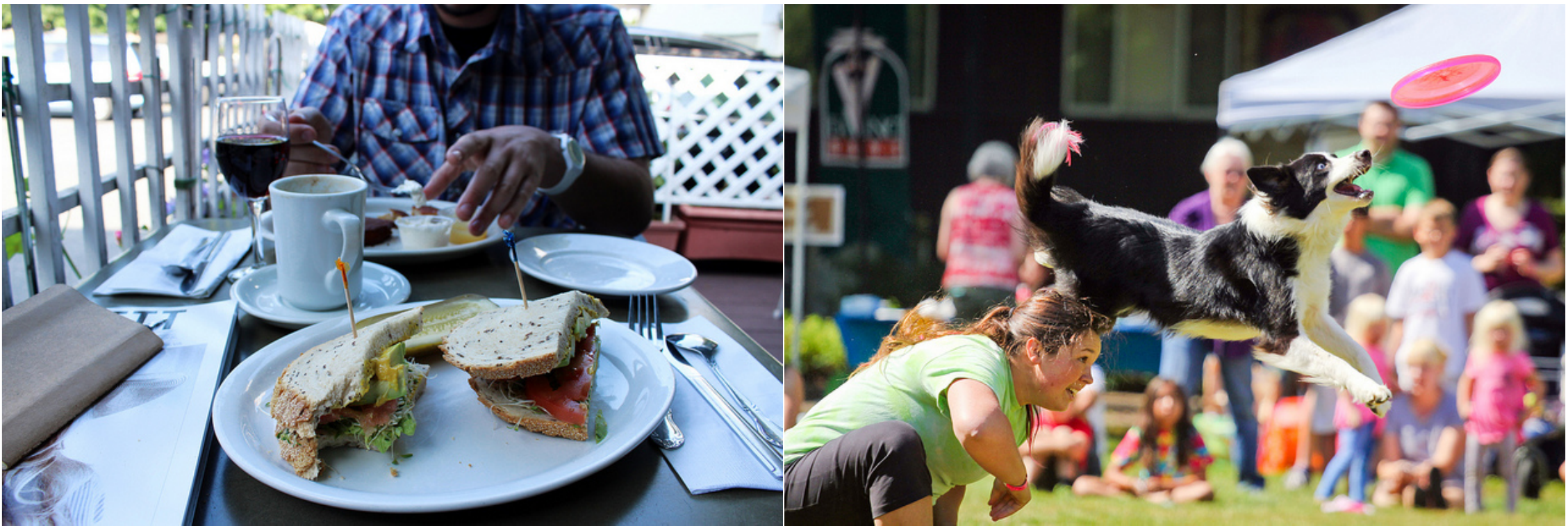}
	\caption{Examples of how contextual layout assists attribute prediction in wild. The \textit{person} (on left) and the \textit{dog} (on right) should be respectively labeled with the attributes \textit{eating} and \textit{catching}. This is hard to agree upon if we would have taken these object instances in isolation, out of their contexts \textit{i.e} food and frisbee.}\label{fig:teaser_context}
\end{figure}

Attributes are different from category labels in three major aspects. \textbf{First}, category labels are agnostic with respect to the intra-class variations that exist among different instances of a single category. Such flat representation cannot distinguish between a \textit{grumpy} cat and a \textit{joyful} one as it only sees them as cats. \textbf{Second}, attributes go across category boundaries. Hence, they can be used to potentially describe an exponential number of object categories (via different combinations) even if the associated category has never been observed before (\textit{e.g} zero-shot learning). \textbf{Third}, unlike category labels that can be effectively inferred from the object itself, humans heavily rely on the contextual cues for the attribute prediction. Take the examples shown in Figure \ref{fig:teaser_context}. If we only consider the bounding box around the dog, one would not assign the attribute \textit{catching} to it. Instead, \textit{running} may even be a valid attribute. However, leveraging contextual layout where the dog is above the ground, and close to a frisbee, provides human with sufficient indications to not only rule out the attribute \textit{running} but also confidently label the dog with the attribute \textit{catching}. Similarly, the table, food and plate, collectively serve as the context, building the ground for associating attribute \textit{eating} to the person.

Considering the aforementioned characteristics of attributes, we hypothesize that the attribute prediction task would benefit from contextual cues if they are properly represented. One can organize the context supervision into three levels: image-level, instance-level and pixel-level. Image-level supervision represents the context as a binary vector indicating whether an instance of a certain category appears somewhere in the context. Therefore, it is blind to the spatial relationships that exist between underlying components \textit{i.e} object instances in the scene. In the instance-level supervision, context is available in terms of a set of category label and bounding box tuples. That is, unlike the image-level, instance-level context supervision can model the spatial relationships in the scene. Lastly, in the pixel-level context supervision, we have access to the category labels in a per-pixel fashion. Obviously, this provides a much stronger supervision signal compared to the other two alternatives. In this work, we propose augmenting attribute prediction by transferring weakly pixel-level context supervision, from an auxiliary semantic segmentation task.

So far, we've explained attributes in general when they describe an instance of an object in a scene. However, the same is valid when attributes characterize variations of a certain object category. In this paper, we are interested in person-related, specifically facial and full body attributes. We view the concept of contextual cues, previously detailed for attributes of objects in the scene, as the natural correspondence of object attributes to the object parts and their associated layout in the spatial domain of the object boundary.

Naturally, attributes are ``additive'' to the objects (\textit{e.g.}, glasses for person). It means that an instance of an object may or may not take a certain attribute, while in either case the category label is preserved (\textit{e.g.}, a person with or without glasses is still labeled as person). Hence, attributes are especially useful in problems that aim at modeling intra-category variations such as fine-grained classification. Despite their additive character, attributes do not appear in arbitrary regions of the objects (\textit{e.g.}, hat if appears, is highly likely to show up on the top of person's head). This notion is the basis of our work. {\em We hypothesize that the attribute prediction can benefit from localization cues}. Specifically, to detect an attribute, instead of processing the entire spatial domain at once, one should focus on the region in which that attribute naturally shows up. However, not all attributes have precise correspondences. For example, it is ambiguous from where in the face, we as humans, infer if a person is \textit{young} or \textit{attractive}. Hence, instead of hard-coding the correspondences, even where those seem clear (\textit{e.g.} glasses with nose and eyes), we allow the model to \textit{learn} how to leverage the localization cues that are transferred from a relevant auxiliary task to the attribute prediction problem.

Using bounding boxes to show the boundary limits of objects is a common practice in computer vision. However, regions that different attributes are associated to drastically vary in terms of appearance. For example, in a face image, one cannot effectively put a bounding box around the region associated to ``hair''. In fact, the shape of the region can be used as an indicative signal on the attribute. On top of that, we have the partial occlusion of object parts which introduces further challenges by arbitrarily deforming visible regions. Therefore, we need an auxiliary task that learns detailed pixel-wise localization information without restricting the corresponding regions to be of certain pre-defined shapes. Semantic segmentation has all the aforementioned characteristics. It is the problem of assigning class labels to every pixel in an image. As a result, a successful semantic segmentation approach has to learn pixel-level localization cues which implicitly encode color, structure, and geometric characteristics in fine detail. In this work, since we are interested in person-related attributes, we take face \cite{smith2013exemplar} and human body \cite{gong2017look} semantic parsing problems as auxiliary tasks to steer the spatial focus of the attribute prediction methods accordingly.

To perform attribute prediction, we feed an image to a fully convolutional neural network which generates feature maps that are ready to be aggregated and passed to the classifier. However, global pooling \cite{lin2013network} is agnostic to where, in spatial domain, the attribute-discriminative activations occur. Hence, instead of propagating the attribute signal to the entire spatial domain, we funnel it into the semantic regions. By doing so, our model learns \textit{where} to attend and \textit{how} to aggregate the feature map activations. We refer to this approach as Semantic Segmentation-based Pooling (SSP), where activations at the end of the attribute prediction pipeline are pooled within different semantic regions.

Alternatively, we can incorporate the semantic regions into earlier layers of the attribute prediction network with a gating mechanism. Specifically, we propose augmenting max pooling operations such that they do not mix activations that reside in different semantic regions. Our approach generates multiple versions of the activation maps that are masked differently and presumably discriminative for various attributes.  We refer to this approach as Semantic Segmentation-based Gating (SSG).

\begin{figure}
	\centering
	\includegraphics[width=0.47\textwidth]{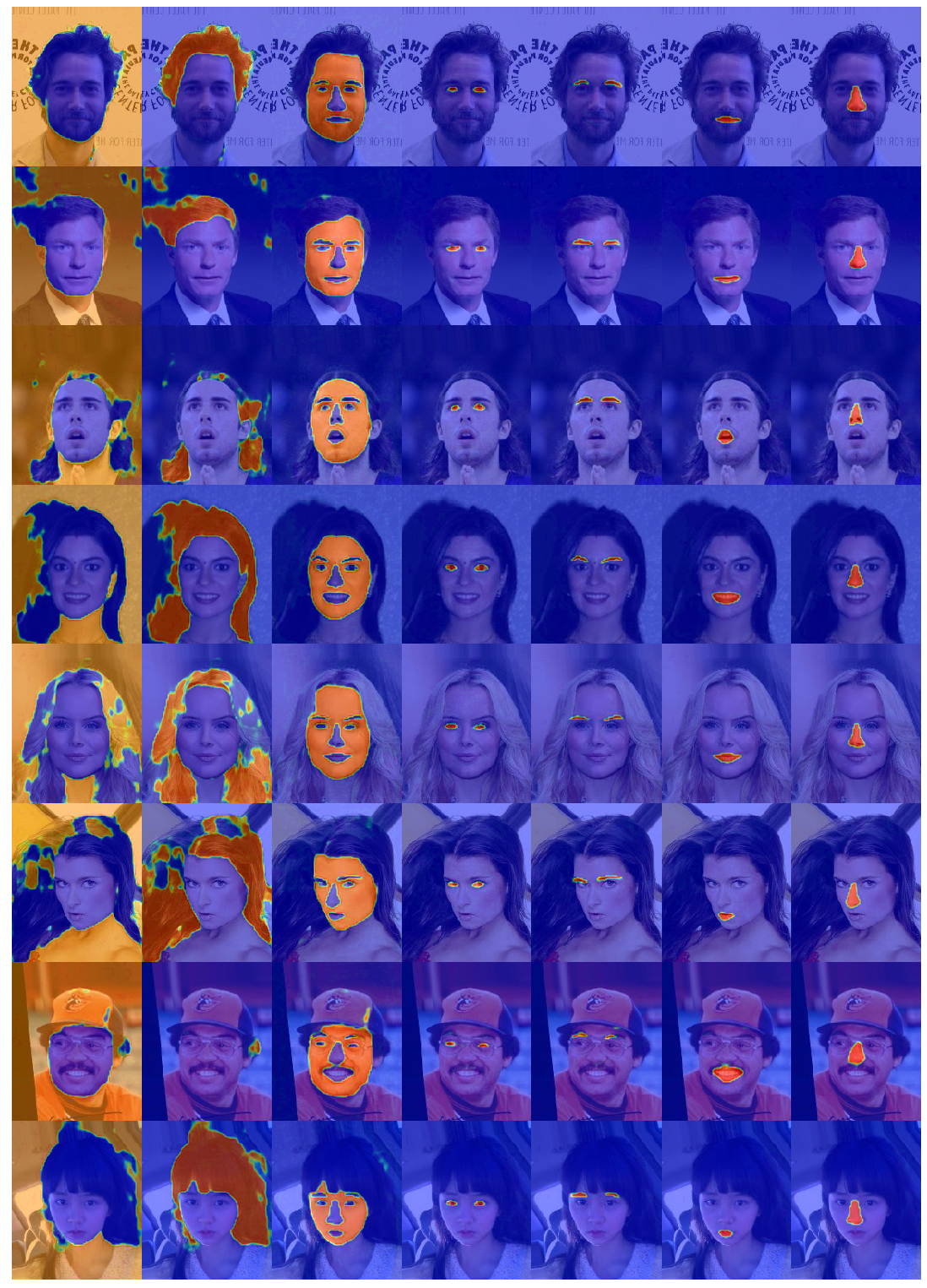}
	\caption{Examples of the segmentation masks generated by our semantic segmentation network \cite{kalayeh2017improving} for previously unseen images. From left to right: background, hair, face skin, eyes, eyebrows, mouth and nose.} \label{fig:segnetexample}
\end{figure}

Since the semantic regions are not available for the attribute benchmarks, we learn to \textit{estimate} them using a deep semantic segmentation network. In our earlier work \cite{kalayeh2017improving}, we took a conceptually similar approach to \cite{noh2015learning} in which an encoder-decoder model was built using convolution and deconvolution layers. However, considering the relatively small number of available data for the auxiliary segmentation task, we had to modify the network architecture. Despite being much simpler than \cite{noh2015learning}, we found our semantic segmentation network \cite{kalayeh2017improving} to be very effective in solving the auxiliary task of semantic face parsing. Examples of the segmentation masks generated for previously unseen images are illustrated in Figure \ref{fig:segnetexample}. Once trained, such network is able to provide localization cues in the form of masks (decoder output) that decompose the spatial domain of an image into mutually exclusive semantic regions. We show that both SSP and SSG mechanisms outperform almost all the existing state-of-the-art facial attribute prediction techniques while employing them together results in further improvements.

One issue with SSP and SSG is their memory utilization. Since both architectures use the output of semantic segmentation to create $N_{S}$ (referring to the number of semantic regions) copies of the previous convolution layer activations. Given limited GPU memory budget, this can restrict the application of these layers when $N_{S}$ grows to large values. Instead, we can circumvent this challenge by learning the proper mask per channel. In contrast to SSP and SSG which mask each and every channel of activations with \textit{all} the $N_{S}$ semantic probability maps, in this paper we propose to learn one mask per channel, as weighted superposition of different semantic probability maps (output of semantic segmentation network). Such workaround that can be simply implemented by a $1\times1$ convolution, adds minimum memory utilization overhead and also allows us to simplify SSP and SSG, yielding a single unified architecture that based on where it is applied in the architecture, can mimic the behavior of SSP and SSG.

Following the recent trend in semantic segmentation, instead of an encoder-decoder as in \cite{kalayeh2017improving}, here we utilize a fully convolutional architecture, specifically Inception-V3\cite{szegedy2016rethinking}. Hence, we can unify attribute prediction and semantic segmentation networks by full weight sharing. As a result, unlike \cite{kalayeh2017improving}, we do not need to pre-train the semantic segmentation network before deploying it in attribute prediction pipeline. Instead, both tasks are learned simultaneously in an end-to-end fashion within a single architecture. We go beyond facial attributes \cite{kalayeh2017improving} and demonstrate the effectiveness of employing semantic segmentation in person-related attributes on multiple benchmarks. Finally, we provide comprehensive quantitative evaluation for the case where attributes are jointly trained with semantic segmentation with the aim to boost the latter task. 

In summary, the contributions of this work are as follows:

\begin{itemize}
    \item We demonstrate the effectiveness of employing semantic segmentation to improve person-related attribute prediction.
    \item We propose a simple alternative to Semantic Segmentation-based Pooling and Semantic Segmentation-based Gating with focus on minimum memory utilization overhead.
    \item We unify semantic segmentation and attribute prediction through multi-tasking a single network and training it in an end-to-end fashion.
    \item We achieve state-of-the-art results in person-related attribute prediction on CelebA, LFWA, WIDER Attributes, and Berkeley Attributes of People datasets.
    \item We provide comprehensive experiments, detailing how to improve semantic segmentation task by leveraging image-level attribute annotations.
\end{itemize}

The remainder of this paper is organized as follows. Section \ref{sec:related_work} offers a detailed review of attribute prediction and semantic segmentation literature. In Section \ref{sec:methodology}, we propose semantic segmentation-based pooling and gating, followed by a simple unifying view of them which benefits from considerably lighter memory footprint. We end this section by providing details of our architectures. Experimental results are shown in Section \ref{sec:experimentalresults}. This includes evaluation of facial and person attributes on four datasets, alongside with comprehensive experiments on the effectiveness of leveraging image-level facial attribute annotations to boost semantic face parsing. Finally, we conclude this paper in Section \ref{sec:conclusion}.

\section{Related Work}\label{sec:related_work}
\subsection{Attribute Prediction}\label{sec:attribute_prediction}
Early works in modeling attributes \cite{farhadi2009describing}\cite{lampert2009learning}\cite{ farhadi2010attribute} came around with the intention to change the recognition paradigm from naming objects to describing them. Therefore, instead of directly learning the object categories, one begins with learning a set of attributes that are shared among different categories. Object recognition can then be built upon the attribute scores. Hence, novel categories are seamlessly integrated, via attributes, with previously observed ones. This can be used to ameliorate label misalignment between train and test data. 

Considering the importance of human category, research in person-related attribute prediction \cite{kumar2009attribute}\cite{kumar2008facetracer}\cite{liu2015faceattributes}\cite{liu2011recognizing}\cite{berg2013poof}\cite{bourdev2011describing} has flourished over the years. To perform attribute prediction, some of the previous works have invested in modeling the correlation among attributes \cite{chen2012describing}\cite{hwang2011sharing}\cite{ jayaraman2014decorrelating}\cite{vedaldi2014understanding}, while others have focused on leveraging the category information \cite{wang2010discriminative}\cite{parikh2011interactively}\cite{gan2016learning}. There are also efforts to exploit the context \cite{li2016human}. 

Another way to view the attribute prediction literature is to divide it into holistic versus part-based methods. The common theme among the holistic models is to take the entire spatial domain into account when extracting features from images. On the other hand, part-based methods begin with an attribute-related part detection and then use the located parts, in isolation from the rest of spatial domain, to extract features. It has been shown that part-based models generally outperform the holistic methods. However, they are prone to the localization error as it can affect the quality of the extracted features. Although, there are works that have taken a hybrid approach benefiting from both the holistic and part-based cues \cite{gkioxari2015actions}\cite{liu2015faceattributes}. Our proposed methods fall in between the two ends of the spectrum. While we process the image in a holistic fashion, we employ localization cues in form of pixel-level semantic representations.

Among earlier works we refer to \cite{kumar2009attribute}\cite{berg2013poof}\cite{bourdev2011describing}\cite{zhang2014panda} as successful examples of part-based attribute prediction models. More recently, in an effort to combine part-based models with deep learning, Zhang \textit{et al.} \cite{zhang2014panda} proposed PANDA, a pose-normalized convolutional neural network (CNN) to infer human attributes from images. PANDA employs poselets \cite{bourdev2011describing} to localize body parts and then extracts CNN features from the located regions. These features are later used to train SVM classifiers for attribute prediction. Inspired by \cite{zhang2014panda}, while seeking to also leverage the holistic cues, Gkioxari \textit{et al.} \cite{gkioxari2015actions} proposed a unified framework that benefits from both holistic and part-based models through utilizing a deep version of poselets \cite{bourdev2011describing} as part detectors. Liu \textit{et al.} \cite{liu2015faceattributes} have taken a relatively different approach. They show that pre-training on massive number of object categories and then fine-tuning on image level attributes is sufficiently effective in localizing the entire face region. Such weakly supervised method provides them with a localized region where they perform facial attribute prediction. In another part-based approach, Singh \textit{et al.} \cite{singh2016end} use spatial transformer networks \cite{jaderberg2015spatial} to localize the most relevant region associated to a given attribute. They encode such localization cue in a Siamese architecture to perform localization and ranking for relative attributes. Rudd \textit{et al.} \cite{rudd2016moon} have addressed the widely recognized data imbalance issue in attribute prediction, by introducing mixed objective optimization network (MOON). The proposed loss function mixes multiple task objectives with domain adaptive re-weighting of propagated loss. \cite{huang2016learning} and \cite{dong2017class} are more examples of recent works that have tried similarly to address the class imbalance in the multi-label problem of attribute prediction. Li \textit{et al.} have recently proposed lAndmark Free Face AttrIbute pRediction (AFFAIR) \cite{li2018landmark}, a hierarchy of spatial transformation networks that initially crop and align the face region from the entire \textemdash assumed to be in the wild \textemdash input image and then localize relevant parts associated with different attributes. Separate neural network architectures then extract feature representations from global and part-based regions where their fusion is used to predict different facial attributes.

In our earlier work \cite{kalayeh2017improving}, we proposed employing semantic segmentation to capture local characteristics for facial attribute prediction. We utilized semantic masks, obtained from a separate pre-trained semantic segmentation network, to gate and pool the activations, respectively at middle and the end of the attribute prediction architecture. In this journal version of the paper, we extend and improve the proposed framework in \cite{kalayeh2017improving} beyond face, and to the human body within the context of person-related attribute prediction. Our driving force in obtaining local cues is semantic parsing of face and human body. Meanwhile, unlike \cite{kalayeh2017improving} that uses two separate networks for the main and auxiliary tasks, here we employ a heavy weight sharing strategy, unifying the semantic segmentation and attribute prediction architectures into one. Next, we discuss the semantic segmentation literature. 

\subsection{Semantic Segmentation}\label{sec:semantic_segmentation}
Semantic segmentation can be seen as a dense pixel-level multi-class classification problem, where the spatial (spatio-temporal) domain of images (videos) is partitioned using fine contours (volumes) into clusters of pixels (voxels) with homogeneous class labels. Prior to the wide-spread popularity of deep convolutional neural networks (CNN), semantic segmentation used to be solved via traditional classifiers such as Support Vector Machine (SVM) or Random Forest applied to the super-pixels \cite{shotton2008semantic}\cite{shotton2013real}. Conditional Random Field (CRF) was often used in these methods as the post processing technique to smooth the segmentation results, based on the assumption that pixels which fall within a certain vicinity, with similar color intensity, tend to be associated with the same class labels.

Among earlier efforts in using deep convolutional neural networks for semantic segmentation, we can refer to Ciresan \textit{et. al} \cite{ciresan2012deep} work
on automatic segmentation of neuronal structures in electron microscopy images. Although, since the number of classes was limited to only membrane and non-membrane, their problem in fact reduces to foreground detection task. Later, upon tremendous success of deep convolutional neural networks in image classification, researchers began designing semantic segmentation models on the top of CNN models, which were previously trained for other tasks, mainly image classification \cite{long2015fully}\cite{badrinarayanan2017segnet}\cite{yu2015multi}\cite{chen2018deeplab}\cite{liang2015semantic}. These methods, by leveraging supervised pre-training on strongly correlated tasks (\textit{e.g.} often labels in two tasks have some overlap), were able to facilitate training procedure for semantic segmentation. However, such an adoption introduces its very own challenges.

Unlike image classification where the activations just before the classifier are flattened via fully connected layer or global average pooling, semantic segmentation task requires the spatial domain to be maintained, specifically the output segmentation maps should be at least of the same size as the input image. Fully Convolutional Networks\cite{long2015fully} popularized CNN architectures for semantic segmentation. Long \textit{et. al} \cite{long2015fully} proposed transforming fully connected layers into convolution layers along with up-sampling intermediate and final activations, whose spatial domain have reduced due to pooling layers through the network architecture. These techniques enable a classification model to output segmentation maps of arbitrary size when operating on input images of any size. Almost all the subsequent state-of-the-art semantic segmentation methods adopted this paradigm. The performance of semantic segmentation task will be compromised if the spatial information is not well preserved through the network architecture. In contrast, architectures designed for image classification very often use pooling layers to aggregate the context activations while discarding the precise spatial coordinates. To alleviate this conceptual discrepancy, two different classes of architectures have evolved. 

First is the encoder-decoder based approach \cite{noh2015learning} in which the encoder gradually reduces the spatial domain through successive convolution and pooling layers, to generate the bottleneck representation. Then the decoder recovers the spatial domain by applying multiple layers of deconvolution or convolution followed by up-sampling, to the aforementioned bottleneck representation. There are usually shortcut connections from the encoder to the decoder, leveraging details at multiple scales, in order to help decoder recovering fine characteristics more accurately. U-Net\cite{ronneberger2015u} SegNet\cite{badrinarayanan2017segnet}, and RefineNet\cite{lin2016refinenet} are the popular architectures from this class.

The second class of architectures developed around the idea of Dilated or Atrous convolutions \cite{yu2015multi}. Specifically, one can avoid using pooling layers in order to preserve detailed spatial information, but this will dramatically increase the computation cost as the following layers must operate on larger activation maps. However, using Atrous convolution \cite{yu2015multi} with dilation rate equal to the stride of the avoided pooling layer, results in the exact same number of operations as the regular convolution operating on pooled activations\footnote{It is worth pointing out that while the computation cost remains the same, employing dilated convolution demands more memory since the size of activation maps remains intact.}. In other words, dilated or Atrous convolution layer allows for an exponential increase in effective receptive field without reducing the spatial resolution.
In a series of works \cite{chen2017rethinking}\cite{chen2018deeplab}, Chen \textit{et. al.} demonstrated how Atrous convolution and its multi-scale variation, namely Atrous spatial pyramid pooling (ASPP) can be utilized within the framework of fully convolutional neural networks to improve the performance of the semantic segmentation task. While in earlier efforts \cite{chen2018deeplab}, Dense CRF \cite{liang2015semantic} has been used, more recent works \cite{chen2017rethinking} have shown comparable results without using such post-processing technique.

Semantic segmentation can be applied at a finer granularity where instead of the entire scene, an object is semantically parsed into its parts. Among popular examples, readers are encouraged to refer to \cite{smith2013exemplar}\cite{le2012interactive}\cite{kae2013augmenting}\cite{liu2015multi} for face, \cite{chen_cvpr14}\cite{wang2015joint}\cite{hariharan2015hypercolumns}\cite{liang2016semantic}\cite{chen2016attention}\cite{xia2016zoom} for general objects, and \cite{gong2017look}\cite{yamaguchi2012parsing}\cite{liang2015deep}\cite{liang2015human}\cite{dong2013deformable}\cite{liu2015fashion}\cite{yang2014clothing}\cite{yamaguchi2013paper}\cite{liu2015matching} for human body and clothing semantic parsing. 

In this work, since we are interested in attributes describing human, when alluding to semantic segmentation, we specifically mean face and human body semantic parsing. Our semantic segmentation model is a fully convolutional neural network based on Inception-V3 \cite{szegedy2016rethinking} architecture, where following \cite{chen2018deeplab}\cite{chen2017rethinking} we have also incorporated Atrous spatial pyramid pooling (ASPP). In addition to utilizing semantic parsing for person-related attribute prediction, we will provide results on semantic face parsing as well. We show that, training an attribute prediction network with image-level supervision can effectively serve as an initialization for semantic parsing task, when the the number of training instances is limited.

\section{Methodology}\label{sec:methodology}
The underlying idea of this work is to exploit semantic segmentation in order to improve person-related attribute prediction. To do so, we first revisit semantic segmentation-based pooling (SSP) and gating (SSG), initially proposed in our earlier work \cite{kalayeh2017improving}. Then, we propose a considerably simpler architecture, which unifies SSP and SSG designs while approximately mimicking their behavior with drastically smaller memory footprint. Furthermore, unlike \cite{kalayeh2017improving}, where there were two networks, one for semantic segmentation and the other for attribute prediction, here we unify two networks with fully sharing the weights among two tasks, and train in an end-to-end fashion. Note that in \cite{kalayeh2017improving}, once trained independently, the semantic segmentation network was frozen during the attribute prediction task. Moving towards more modern architectures than those used earlier in \cite{kalayeh2017improving}, we describe our new models based on modern Inception-V3 \cite{szegedy2016rethinking} as their backbone. This choice will allow us to further push performance boundaries in person-related attribute prediction task.

\subsection{SSP: Semantic Segmentation-based Pooling}
We argue that attributes usually have a natural correspondence to certain regions within the object boundary. Hence, aggregating the visual information from the entire spatial domain of an image would not capture this property. This is the case for the global average pooling used in our baseline as it is agnostic to where, in the spatial domain, activations occur. Instead of pooling from the entire activation map, we propose to first decompose the activations of the last convolution layer into different semantic regions and then aggregate only those that reside in the same region. Hence, rather than a single vector representation, we obtain multiple features, each representing only one semantic region. This approach has an interesting intuition behind it. In fact, SSP funnels the back-propagation of the label signals, via multiple paths, associated with different semantic regions, through the entire network. This is in contrast with global average pooling that rather equally affects different locations in the spatial domain. We later explore this by visualizing the activation maps of the final convolution layer. 

\begin{figure}
	\centering
	\includegraphics[width=0.47\textwidth]{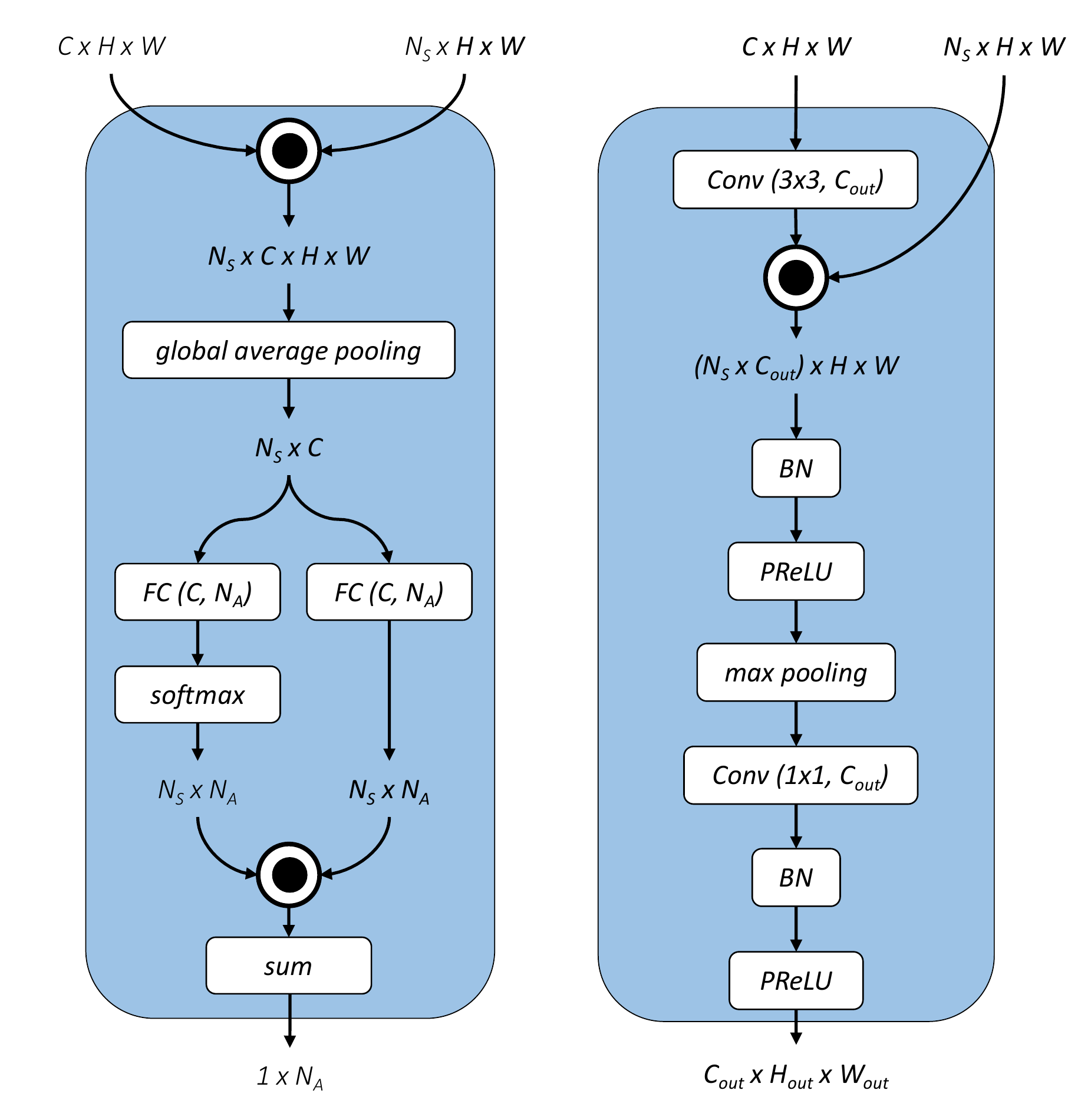}
	\caption{Left: Semantic segmentation-based Pooling (SSP). Right: Semantic segmentation-based Gating (SSG). $N_{S}$ and $N_{A}$, respectively, indicate the number of labels in semantic segmentation and attribute prediction tasks. We assume that the output tensor of activations from the previous layer to either SSP or SSG is of shape $C\times H\times W$ where $C$, $H$ and $W$, respectively represent the number of channels, height and width of the activations. \hlcyan{Alternatively, in Sec. \ref{subsec:sa}, we will show that instead of using all $N_{S}$ semantic regions for every channel, one can learn a single semantic mask per channel. This would also unify the SSP and SSG architectures.}}\label{fig:architecures}
\end{figure}

We can simply concatenate the representations associated with different regions and pass it to the classifier; however, it is interesting to observe if attributes indeed prefer one semantic region to another. Also, whether what our model learns matches human expectation on what attribute corresponds to which region. To do so, we take a similar approach to \cite{Bilen16} where Bilen and Vedaldi employed a two branch network for weakly supervised object detection. We pass the vector representations, each associated with a different semantic region, to two branches one for recognition and another for localization. We implement these branches as linear classifiers that map vector representations to the number of attributes. Hence, we have multiple detection scores for an attribute each inferred based on one and only one semantic region. To combine these detection scores, we normalize outputs of the localization branch using softmax non-linearity across different semantic regions. This is a per-attribute operation, not an across-attribute one. We then compute the final attribute detection scores by a weighted sum of the per-region logits (\textit{i.e.} outputs of recognition branch) using weights generated by the localization branch. Figure \ref{fig:architecures} (Left)  shows the SSP architecture.  

\subsection{SSG: Semantic Segmentation-based Gating}
Max pooling is used to compress the visual information in the activation maps of the convolution layers. Its efficacy has been proven in many computer vision tasks, such as image classification and object detection. However, attribute prediction is inherently different from image classification. In image classification, we want to aggregate the visual information across the entire spatial domain to come up with a single label for the image. In contrast, many attributes are inherently localized to specific image regions. Consequently, aggregating activations that reside in the ``mouth" region with the ones that correspond to ``hair'', would confuse the model in detecting ``smiling" and ``wavy hair" attributes. We propose SSG to cope with this challenge. 

Figure \ref{fig:architecures} (Right),  shows our proposed SSG architecture where $C_{out}$ may or may not be the same as $C$ (similarly for $H$ and $W$). To gate the output activations of the convolution layer, we broadcast element-wise multiplication for each of the semantic regions with the entire activation maps. This generates multiple copies of the activations that are masked differently. In other words, such mechanism spatially decomposes the activations into copies, where large values cannot simultaneously occur in two semantically different regions. For example, gating with the semantic mask that corresponds to the ``mouth'' region, would suppress the activations falling outside its area while preserving those that reside inside it. However, the area which a semantic region occupies varies from one image to another. 

We observed that, directly applying the output of the semantic segmentation network results in instabilities in the middle of the network. To alleviate this, prior to the gating procedure, we normalize the semantic masks such that the values of each channel sums up to 1. We then gate the activations right after the convolution and before the batch normalization \cite{ioffe2015batch}. This is very important since the batch normalization \cite{ioffe2015batch} enforces a normal distribution on the output of the gating procedure. Then, we can apply max pooling on these gated activation maps. Since, given a channel, activations can only occur within a single semantic region, max pooling operation cannot blend activation values that reside in different semantic regions. We later restore the number of channels using a $1\times1$ convolution. It is worth noting that SSG can potentially mimic the standard max pooling by learning  a sparse set of weights for the $1\times1$ convolution. In a nutshell, semantic segmentation-based gating allows us to process the activations of convolution layers in a per-semantic region fashion while it also learns how to blend the pooled values back in.

\begin{figure*}
	\centering
	\includegraphics[width=1.\textwidth]{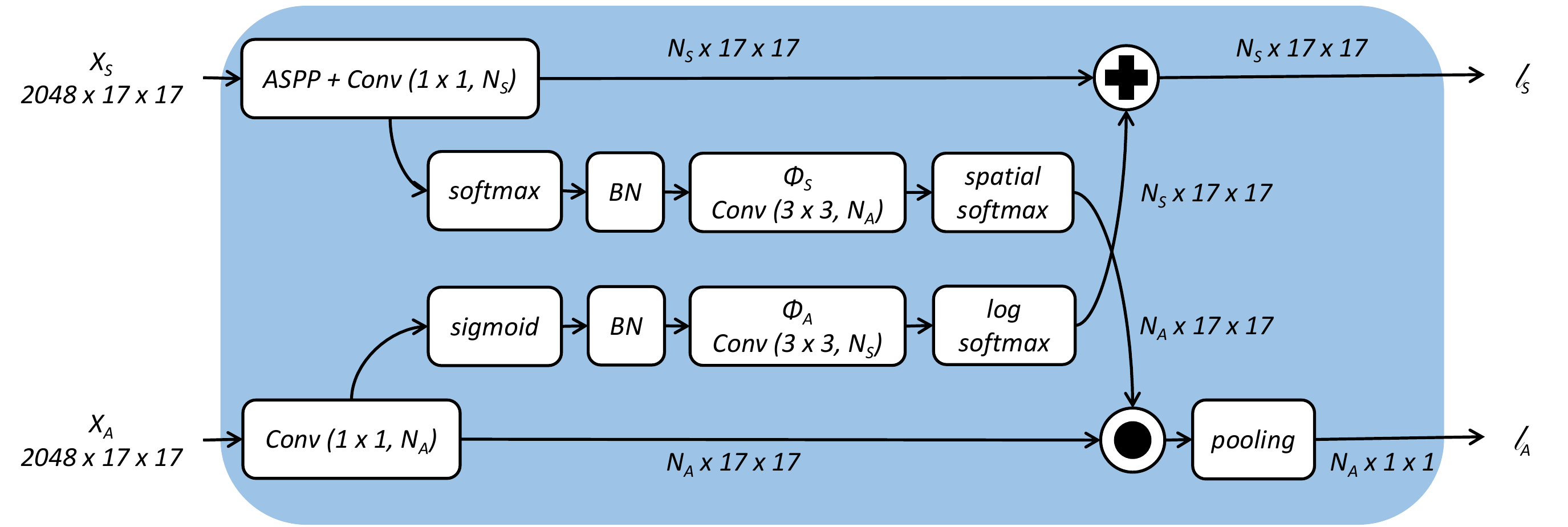}
	\caption{\hlcyan{Architecture of the Symbiotic Augmentation (SA). The embedding layers, $\Phi_{S}$ and $\Phi_{A}$, respectively utilize the output of semantic segmentation and attribute prediction classifiers to augment the other task. Similar to Figure \ref{fig:architecures}, $N_{S}$ and $N_{A}$ denote the number of output labels for semantic segmentation and attribute prediction, where, $l_{S}$ and $l_{A}$ are their corresponding loss functions (per-pixel softmax cross entropy, image-level sigmoid cross entropy). Addition and multiplication are element-wise operations.}}\label{fig:improved_ssg}
\end{figure*}

\subsection{A Simple Unified View to SSP and SSG}\label{subsec:sa}
In both SSP and SSG architectures, we use the output of semantic segmentation to create $N_{S}$ copies of the activations. Each copy, assuming semantic parsing outputs are perfect, preserves the activation values residing in one semantic region while suppressing those that are outside that. Hence, both SSP and SSG should maintain $N_{S}$ times the size of activation maps in the memory. As $N_{S}$ value grows, this can certainly become problematic due to limited GPU memory budget. A simple workaround for this is to learn the masks per channel. Specifically, instead of masking each and every channels of the previous convolution activations by \textit{all} the $N_{S}$ semantic probability maps, we learn one mask per channel (ref. $\Phi_{S}$ in Figure \ref{fig:improved_ssg}). This can be simply implemented via a $1\times1$ convolution on the top of semantic segmentation probability maps. However, in practice, we observed that larger kernels can result in slight performance gain. Similar to SSG, the output logits of the semantic segmentation classifier must be normalized, via batch normalization, prior to being passed to the embedding convolution layer. The output of the embedding should also be spatially normalized. Such embedding allows the model to either pick one or combine (weighted superposition) multiple semantic maps, in order to generate proper mask for each channel. We initialize the convolution kernels $\Phi_{S}$ of the embedding layers with zeros and no bias. This is inspired by the idea that each channel should start by using all the semantic regions equally. However, through training, it has the freedom to change towards combining only a selected number of regions. We later visualize how the learned convolution kernels of $\Phi_{S}$ look like in Figures \ref{fig:weights_face_S} and \ref{fig:weights_body_S}. 

We now go one step further as the same idea can be used when we reverse the roles of tasks. In particular, we can use the output of attribute prediction to guide the semantic segmentation task. We refer to this joint semantic augmenting model, illustrated in Figure \ref{fig:improved_ssg}, as Symbiotic Augmentation (SA). The architecture of the embedding module in this case, $\Phi_A$, is the same as $\Phi_S$ except the normalization operations are done differently. Figure \ref{fig:improved_ssg} shows that in Symbiotic Augmentation, each task augments the other task's representation, through its corresponding output logits, while simultaneously being trained in an end-to-end fashion. This is different than SSP and SSG, where only a pre-trained semantic segmentation model, while frozen at deployment, augments attribute prediction task. Note that, in addition to a lower memory footprint\footnote{\hlcyan{The memory footprint of SSP is of $\mathcal{O}(N_{S}CHW) + \mathcal{O}(N_{S}N_{A})$ while SA's is of $\mathcal{O}(N_{S}HW)+\mathcal{O}(N_{A}HW)$. Here $C$ refers to the number of output channels in last (before classifier) convolution layer, while $H$ and $W$ respectively denote height and width of the final spatial resolution.}}, this approach allows us to simplify the SSP by unifying the recognition and localization branches. That is because the learned masks can properly weigh each channel and the order of consecutive linear operations (matrix multiplication through fully connected layer and scaling through weights of localization branch) is interchangeable. 

\subsection{Network Architectures}
\begin{figure*}
	\centering
	\includegraphics[width=1.\textwidth]{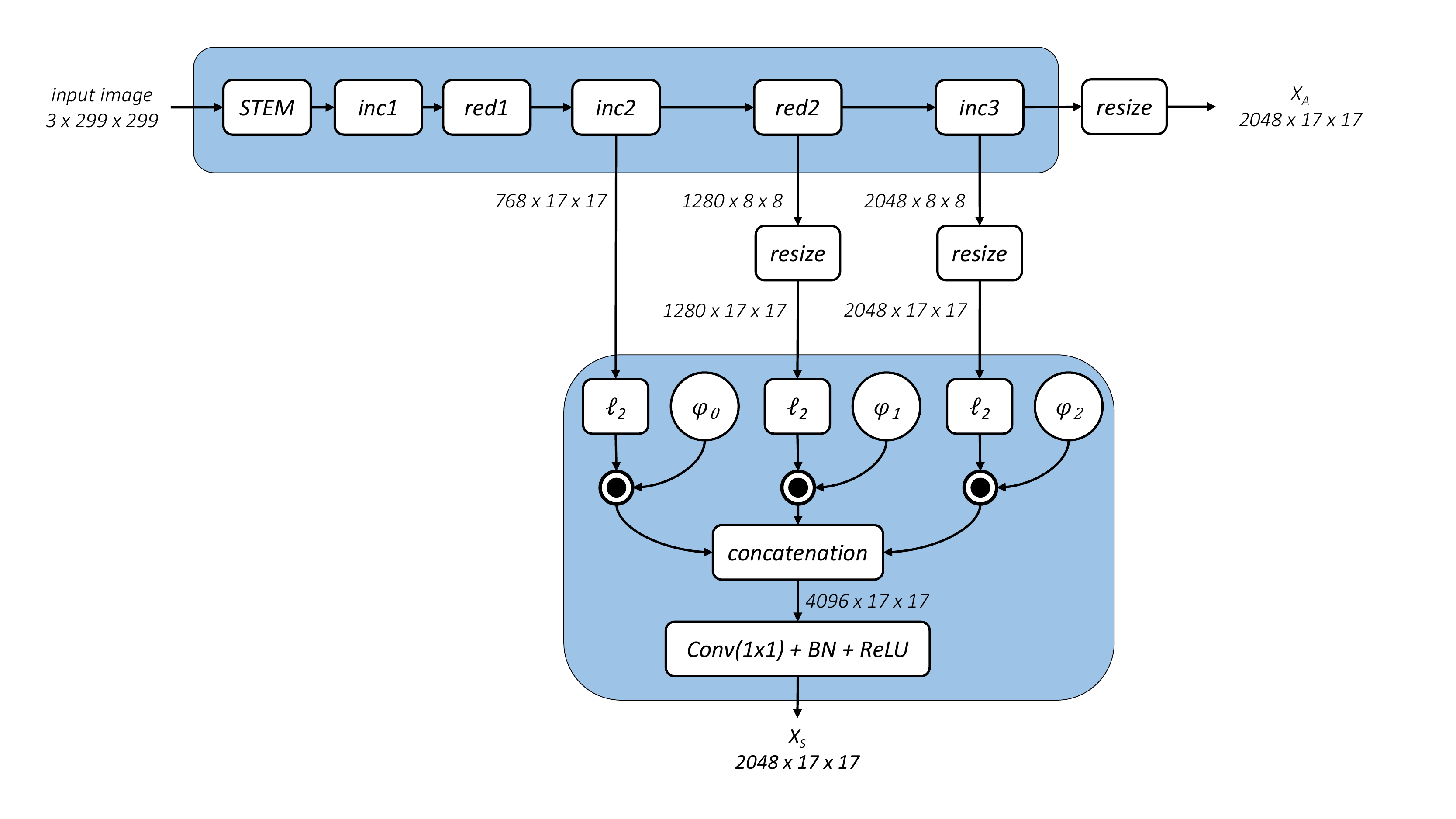}
	\caption{\hlcyan{Inception-V3 \cite{szegedy2016rethinking} backbone architecture used in the Symbiotic Augmentation (SA) experiments. $X_{A}$ and $X_{S}$ are used as input features to SA (ref. Figure \ref{fig:improved_ssg}). In order to generate $X_{S}$, we $\ell_{2}$ normalize the intermediate activations and scale them by learnable $\varphi_*$ parameters. Refer to \cite{szegedy2016rethinking} for the details of the Inception-V3 architecture.}}\label{fig:weightsharing}
\end{figure*}

We use Inception-V3 \cite{szegedy2016rethinking} as the convolutional backbone of Symbiotic Augmentation (SA), for both semantic segmentation and attribute prediction models. Its architecture is 48 layers deep and uses global average pooling instead of fully-connected layers which allows operating on arbitrary input image sizes. Inception-V3 \cite{szegedy2016rethinking} has a total output stride of 32. However, to maintain low computation cost and memory utilization, the size of activation maps quickly reduces by a factor of 8 in only first seven layers, referred to as STEM \cite{szegedy2016rethinking} in Figure \ref{fig:weightsharing}. This is done by one convolution and two max pooling layers that operate with the stride of 2. The network follows by three blocks of Inception layers separated by two grid reduction modules. Spatial resolution of the activations remains intact within the Inception blocks, while grid reduction modules halve the activation size and increase the number of channels. For more details on the Inception-V3 \cite{szegedy2016rethinking} architecture, readers are encouraged to refer to \cite{szegedy2016rethinking}. Note that, for SSP, SSG and SSP+SSG experiments which were initially reported in \cite{kalayeh2017improving}, a VGG16-like backbone architecture has been used. Further details are provided in \cite{kalayeh2017improving}.

In this work, we use a single architecture to simultaneously learn semantic parsing and attribute prediction tasks. This is different than \cite{kalayeh2017improving} where semantic segmentation model was pre-trained and then deployed (weights were frozen) into attribute prediction pipeline. Specifically, we share the weights of the Inception-V3 \cite{szegedy2016rethinking} while training with a mixed minibatch that is comprised of equal instances associated to attribute prediction and semantic segmentation tasks. Figure \ref{fig:weightsharing} illustrates how we obtain feature representations for both tasks using a single architecture. Note that each element in the minibatch has only one type of annotations, either attribute or semantic segmentation labels. Hence, when $X_{A}$ and $X_{S}$ are passed to the Symbiotic Augmentation (SA), shown in Figure \ref{fig:improved_ssg}, depending on the annotation type, either $l_S$ or $l_A$ is calculated.

\section{Experiments}\label{sec:experimentalresults}

\subsection{Datasets and Evaluation Measures}

We evaluate our proposed attribute prediction models on multiple benchmarks. Specifically, we use CelebA and LFWA \cite{liu2015faceattributes} for facial attributes, while benchmarking on WIDER Attribute \cite{li2016human} and Berkeley
Attributes of People \cite{bourdev2011describing} for person attribute prediction. 

Liu \textit{et al.} \cite{liu2015faceattributes} have used classification accuracy/error as the evaluation measure on CelebA and LFWA. However, we believe that due to significant imbalance between the numbers of positive and negatives instances per attribute, such measure cannot appropriately evaluate the quality of different methods. Similar point has been raised by \cite{rudd2016moon, huang2016learning, dong2017class} as well. Therefore, in addition to the classification error, we also report the average precision (AP) of the prediction scores. Following the literature, we solely report AP for WIDER Attribute \cite{li2016human} and Berkeley Attributes of People \cite{bourdev2011describing}. Since attribute benchmarks do not come with pixel-level labels, we train our semantic segmentation model on auxiliary datasets. For experiments corresponding to facial attributes, we use Helen Face \cite{le2012interactive} along with segment label annotations supplemented by \cite{smith2013exemplar}. For person attribute prediction experiments, we train the semantic parsing model on Look into Person (LIP) \cite{gong2017look} dataset. We use the standard data split of each corresponding dataset. 

%Following \cite{gong2017look}, we report the performance on LIP using mean IoU, mean class accuracy and overall pixel accuracy. However, the quality of segmentation on Helen Face \cite{le2012interactive}, as reported in \cite{smith2013exemplar} and subsequent works, will be measured by F score \Mahdi{numbers are currently in miou. So either remove this or compute F score for the corresponding experiments}. Next, we provide details on each of the aforementioned datasets. 

\textbf{CelebA} \cite{liu2015faceattributes} consists of 202,599 images partitioned into training, validation and test splits with approximately 162K, 20K and 20K images in the respective splits. There are a total of 10K identities (20 images per identity) with no identity overlap between evaluation splits. However, we do not use identity annotations. Images are annotated with 40 facial attributes such as, ``wavy hair", ``mouth slightly open", and ``big lips". In addition to the original images, CelebA provides a set of pre-cropped images. We report our results on both of these image sets. 

\textbf{LFWA} \cite{liu2015faceattributes} has a total of 13,143 images of 5,749 identities with pre-defined train and test splits, which divide the entire dataset into two approximately equal partitions. Each image is annotated with the same 40 attributes used in CelebA\cite{liu2015faceattributes}.

\textbf{WIDER Attribute} \cite{li2016human} is collected from 13,789 WIDER images \cite{xiong2015recognize}, containing usually many people in each image with huge human variations. Each person in these images is then annotated with a bounding box and 14 distinct human attributes such as ``longhair'', ``sunglasses'', ``hat'', ``skirt'', and ``facemask''. This results in a total of 57,524 boxes. Out of 13,789 images, WIDER Attribute \cite{li2016human} is split into 5,509 training, 1,362 validation and 6,918 test images. There are 30 scene-level labels that each image is annotated with. However, we do not use them and solely train and evaluate on bounding boxes of people. We evaluate on the 29,179 bounding boxes from testing images, after training on 28,345 person boxes extracted from aggregation of training and validation images. Unlike CelebA and LFWA \cite{liu2015faceattributes}, missing attributes are allowed in WIDER Attribute \cite{li2016human} dataset.

\textbf{Berkeley Attributes of People} \cite{bourdev2011describing} contains 4,013 training and 4,022 test instances. The example images are centered at the person and labeled with 9 attributes namely, ``is male'', ``has long hair'', ``has glasses'', ``has hat'', ``has tshirt'', ``has long sleeves'', ``has shorts'', ``has jeans'', ``has long pants''. Similar to the WIDER Attribute \cite{li2016human}, here unspecified attributes are also allowed.

\textbf{Helen Face} \cite{le2012interactive} consists of 2,330 images with highly accurate and consistent annotations of the primary facial components. Smith \textit{et. al} \cite{smith2013exemplar} have supplemented Helen Face \cite{le2012interactive} with 11 segment label \footnote{``background'', ``face skin'' (excluding ears and neck), ``left eyebrow'', ``right eyebrow'', ``left eye'', ``right eye'', ``nose'', ``upper lip'', ``inner mouth'', ``lower lip'' and ``hair''} annotations per image. Images are divided into splits of 2000, 230 and 100, respectively for training, validation and test. We train our semantic segmentation model on the aggregation of training and validation splits and evaluate on the test split. 
% For facial attribute prediction experiments, once the semantic face parsing model was trained, we combine left and right eye (eyebrow) labels to create a single eye (eyebrow) label. Similarly, we aggregate ``upper lip'', inner mouth, and lower lip to generate a single mouth label. As a result we end up with a total of 7 semantic regions of ``background'', ``hair'', ``face skin'', ``eyes'', ``eyebrows'', ``mouth'' and ``nose''. 

\textbf{LIP} \cite{gong2017look} consists of $\sim$30,000 and 10,000 images respectively for train and validation. Each images is annotated with 20 semantic labels\footnote{``Background'', ``Hat'', ``Hair'', ``Glove'', ``Sunglasses'', ``Upper-clothes'', ``Dress'', ``Coat'', ``Socks'', ``Pants'', ``Jumpsuits'', ``Scarf'', ``Skirt'', ``Face'', ``Right-arm'', ``Left-arm'', ``Right-leg'', ``Left-leg'', ``Right-shoe'' and ``Left-shoe''}. 
% For human attribute prediction, once semantic parsing model was trained, we aggregate output segmentation (probability) maps associated to different relevant labels to create 5 coarse regions specifically, ``Foreground'', ``Head'', ``Upper-body'', ``Lower-body'' and ``Shoes''.

\subsection{Evaluation of Facial Attribute Prediction}
For all the numbers reported here, we want to point out that FaceTracer \cite{kumar2008facetracer} and PANDA \cite{zhang2014panda} use groundtruth landmark points to attain face parts. Wang \textit{et al.} \cite{wang2016walk} use 5 million auxiliary image pairs, collected by the authors, to pre-train their model. Wang \textit{et al.} \cite{wang2016walk} also use state-of-the-art face detection and alignment to extract the face region from CelebA and LFWA images. \textit{However, we train all our models with only attribute and auxiliary face/human parsing labels}. 

% \subsubsection{Evaluation on CelebA}
We compare our proposed method with the existing state-of-the-art attribute prediction techniques on the CelebA \cite{liu2015faceattributes}. To prevent any confusion and have a fair comparison, Table \ref{tab:CelebAresult} reports the performances in two separate columns distinguishing the experiments that are conducted on the original image set from those where the pre-cropped image set have been used. %We see that our base model with global average pooling, thanks to a more modern architecture, not only outperforms our earlier average pooling \cite{kalayeh2017improving} model but also semantic segmentation based ones with the exception of Kalayeh \textit{et. al} \cite{kalayeh2017improving} SSP+SSG model operating on pre-cropped images.

Experimental results indicate that under different settings and evaluation protocols, our proposed semantic segmentation-based pooling and gating mechanisms can be effectively used to boost the facial attribute prediction performance. That is particularly important given that our global average pooling baselines already beat almost all the existing state-of-the-art methods. To see if SSP and SSG are complementary to each other, we also report their combination where the corresponding predictions are simply averaged. We observe that such process further boosts the performance. 

\begin{table}
	\centering
	\begin{tabular}{lcc}
		\toprule
		\multicolumn{3}{c}{\textbf{Classification Error(\%)}}\\
		\midrule
		Method & Original & Pre-cropped\\
		\midrule
		FaceTracer \cite{kumar2008facetracer} & 18.88 & --\\
		PANDA \cite{zhang2014panda} & 15.00 & --\\		
		Liu \textit{et al.} \cite{liu2015faceattributes} & 12.70 & --\\
		Wang \textit{et al.} \cite{wang2016walk} & 12.00 & --\\
		Zhong \textit{et al.} \cite{zhong2016leveraging} & 10.20 & --\\
		Rudd \textit{et al.} \cite{rudd2016moon}: Separate & -- & 9.78\\
		Rudd \textit{et al.} \cite{rudd2016moon}: MOON & -- & 9.06\\
		AFFAIR \cite{li2018landmark} & 8.55 & --\\
		\midrule
		SPPNet$^{*}$ & -- & 9.49\\	
		Naive Approach & 9.62 & 9.13\\
		BBox & -- & 8.76\\
		Avg. Pooling & 9.83 & 9.14\\
		SSG &  9.13 & 8.38\\
		SSP & 8.98 & 8.33\\
		SSP + SSG &  8.84 & \textbf{8.20}\\
		Inception-V3: baseline & 8.68 & --\\
		Symbiotic Augmentation (SA) & \textbf{8.53} & --\\
		\toprule
		\multicolumn{3}{c}{\textbf{Average Precision(\%)}}\\
		\midrule
		Method & Original & Pre-cropped\\
		\midrule
		AFFAIR \cite{li2018landmark} & 79.63 & --\\
		\midrule
		SPPNet$^{*}$ & -- & 77.69\\
		Naive Approach & 76.29 & 79.74\\
		BBox & -- & 79.95\\
		Avg. Pooling & 77.16 & 79.74\\
		SSG & 77.46 & 80.55\\		
		SSP & 78.01 & 81.02\\
		SSP + SSG & 78.74 & \textbf{81.45}\\
		Inception-V3: baseline & 79.28 & --\\
		Symbiotic Augmentation (SA) & \textbf{80.10} & --\\		
		\toprule
		\multicolumn{3}{c}{\textbf{Balanced Accuracy(\%) \cite{huang2016learning}}}\\
		\midrule
		Method & Original & Pre-cropped\\
		\midrule
		Huang \textit{et al.} \cite{huang2016learning} & -- & 84.00\\
		CRL(C) \cite{dong2017class} & -- & 85.00\\
		CRL(I) \cite{dong2017class} & -- & 86.00\\	
		\midrule
		Avg. Pooling & -- & 86.73\\
		SSG & -- & 87.82\\
		SSP & -- & \textbf{88.24}\\
		\bottomrule\\
	\end{tabular}
	\caption{Attribute prediction performance evaluated by the classification error, average precision and balanced classification accuracy \cite{huang2016learning} on the CelebA \cite{liu2015faceattributes} original and pre-cropped image sets.}\label{tab:CelebAresult}
\end{table}

To investigate the importance of aggregating features within the semantic regions, we replace the global average pooling in our basic model with the spatial pyramid pooling layer \cite{he2014spatial}. We use a pyramid of two levels and refer to this baseline as SPPNet$^*$. While aggregating the output activations in different locations, SPPNet$^*$ does not align its pooling regions according to the semantic context that appears in the image. This is in direct contrast with the intuition behind our proposed methods. Experimental results shown in Table \ref{tab:CelebAresult} confirm that simply pooling the output activations at multiple locations is not sufficient. In fact, it results in a lower performance than global average pooling. This verifies that the improvement obtained by our proposed models is due to their content aware pooling/gating mechanisms.

\textbf{Naive Approach }A naive alternative approach is to consider the segmentation maps as additional input channels. To evaluate its effectiveness, we feed the average pooling basic model with 10 input channels, 3 for RGB colors and 7 for different semantic segmentation maps. The input is normalized using Batch Normalization \cite{ioffe2015batch}. We train the network using the same setting as other aforementioned models. Our experimental results indicate that such naive approach cannot leverage the localization cues as good as our proposed methods. Table \ref{tab:CelebAresult} shows that at best, the naive approach is on par with the average pooling basic model. We emphasize that feeding semantic segmentation maps along with RGB color channels to a convolutional network results in blending the two modalities in an \textit{additive} fashion. Instead, our proposed mechanisms take a \textit{multiplicative} approach by masking the activations using the semantic segmentation probability maps.

\textbf{Semantic Masks vs. Bounding Boxes }To analyze the necessity of semantic segmentation, we generate a baseline, namely BBox, which is similar to SSP. However, we replace the semantic masks in SSP with the bounding boxes on the facial landmarks. Note that we use the groundtruth location of the facial landmarks, provided in CelebA dataset \cite{liu2015faceattributes}, to construct the bounding boxes. Hence, to some extent, the performance of BBox is the upper bound of the bounding box experiment. There are 5 facial landmarks including left eye, right eye, nose, left mouth and right mouth. We use boxes with area $20^2$ ($40^2$ gives similar results) and 1:1, 1:2 and 2:1 aspect ratios. Thus, there are a total of 16 regions including the whole image itself. From Table \ref{tab:CelebAresult}, we see that our proposed models, regardless of the evaluation measure, outperform the bounding box alternative, suggesting that semantic masks should be favored over the bounding boxes on the facial landmarks.

\textbf{Balanced Classification Accuracy } Given the significant imbalance in the attribute classes, also noted by \cite{huang2016learning, rudd2016moon, dong2017class}, we suggest using average precision instead of classification accuracy/error to evaluate attribute prediction. Instead, Huang \textit{et al.} \cite{huang2016learning} and later \cite{dong2017class} have adopted balanced accuracy measure. To evaluate our proposed approach in balanced accuracy measure, we fine-tuned our models with the weighted ($\propto$ imbalance level) binary cross entropy loss. From Table \ref{tab:CelebAresult}, we observe that under such measure, all the variations of our proposed model outperform both \cite{huang2016learning} and \cite{dong2017class} with large margins.

\begin{table}
	\centering
	\begin{tabular}{lcc}
		\toprule
		Method & Classification Error(\%)& AP(\%)\\
		\midrule
		FaceTracer \cite{kumar2008facetracer} & 26.00 & --\\
		PANDA \cite{zhang2014panda} & 19.00 & --\\		
		Liu \textit{et al.} \cite{liu2015faceattributes} &  16.00 & --\\
		Zhong \textit{et al.} \cite{zhong2016leveraging} & 14.10 & --\\
		Wang \textit{et al.} \cite{wang2016walk} & 13.00 & --\\
		AFFAIR \cite{li2018landmark} & 13.87 & 83.01\\
		\midrule
		Avg. Pooling & 14.73 & 82.69\\
		SSG &  13.87 & 83.49\\
		SSP & 13.20 & 84.53\\
		SSP + SSG &  \textbf{12.87} & \textbf{85.28}\\
		\bottomrule\\
	\end{tabular}
	\caption{Attribute prediction performance evaluated by the classification error and the average precision (AP) on LFWA \cite{liu2015faceattributes} dataset.}\label{tab:resultsLFWA}
\end{table}

To better understand the effectiveness of our proposed approach on facial attributes, we also report experimental results on the LFWA dataset \cite{liu2015faceattributes} in Table \ref{tab:resultsLFWA}. Here, we observe a similar trend to the one in CelebA, where all the proposed models which exploit localization cues successfully improve the baseline. Specifically, SSP + SSG achieves considerably better performance than the average pooling model with margins of 1.86\% in classification accuracy and 2.59\% in average precision. Our best model also outperforms all other state-of-the-art methods.

\textbf{Symbiotic Augmentation (SA)} All the results reported so far were using a VGG16-like architecture for attribute prediction and a separate pre-trained encoder-decoder architecture for semantic segmentation \cite{kalayeh2017improving}. However, in SA-based models, we have unified the two architectures and train simultaneously with two objective functions. Table \ref{tab:CelebAresult} shows that simply using a stronger convolutional backbone like Inception-V3 \cite{szegedy2016rethinking} boosts the performance on CelebA original image set. Furthermore, SA-based model which is built on the top of such backbone, despite heavily sharing all the weight across two tasks, can achieve even better results, outperforming SSP+SSG and current state-of-the-art AFFAIR \cite{li2018landmark}. However, on LFWA dataset \cite{liu2015faceattributes}, we observed that Inception-V3 \cite{szegedy2016rethinking} baseline performs on par with Avg. Pooling baseline reported in Table \ref{tab:resultsLFWA} and SA cannot obtain a meaningful gain over its counter global average pooling baseline. We also tried (not reported here) solely using LFWA training instances, without pre-training on CelebA, and observed that SA was indeed effective. However it was not able to reach the performance of the model initialized with CelebA. Detailed per-attribute results of our top models for both CelebA and LFWA datasets are shown in Table \ref{tab:per_attr_face}.

\begin{table}
	\centering
	\begin{tabular}{lcc}
		\toprule
		Method & AP(\%)\\
		\midrule
		Fast R-CNN \cite{girshick2015fast} & 80.00\\
		R*CNN \cite{gkioxari2015contextual} & 80.50\\
		Deep Hierarchical Contexts \cite{li2016human} & 81.30 \\
		VeSPA \cite{sarfraz2017deep} & 82.40\\
		ResNet-101 \cite{zhu2017learning} & 85.00\\
		ResNet-SRN-att \cite{zhu2017learning} & 85.40\\
		ResNet-SRN \cite{zhu2017learning} & 86.20\\
		Sarafianos \textit{et. al.} \cite{Sarafianos_2018_ECCV} & 86.40\\
		\midrule
		Inception-V3: baseline & 85.86\\
		Symbiotic Augmentation (SA) & \textbf{87.58}\\
		\bottomrule\\
	\end{tabular}
	\caption{Attribute prediction performance evaluated by the average precision(\%) on WIDER Attribute \cite{li2016human} dataset.}\label{tab:resultsWIDER}
\end{table}

\begin{table}
	\centering
	\begin{tabular}{lcc}
		\toprule
		Method & AP(\%)\\
		\midrule
		Fast R-CNN \cite{girshick2015fast} & 87.80\\
		R*CNN \cite{gkioxari2015contextual} & 89.20\\
		Gkioxari \textit{et al.} \cite{gkioxari2015actions} & 89.50\\
		Deep Hierarchical Contexts \cite{li2016human} & 92.20 \\
		\midrule
		Inception-V3: baseline & 92.87\\
		Symbiotic Augmentation (SA) & \textbf{94.80}\\
		\bottomrule\\
	\end{tabular}
	\caption{Attribute prediction performance evaluated by the average precision(\%) on Berkeley Attributes of People \cite{bourdev2011describing} dataset.}\label{tab:resultsBAPD}
\end{table}

\subsection{Evaluation of Person Attribute Prediction}
Table \ref{tab:resultsWIDER} compares our proposed method with the state-of-the-art on WIDER Attribute \cite{li2016human} dataset. We observe that the Inception-V3 \cite{szegedy2016rethinking} baseline, despite being considerably shallower, performs on par with ResNet-101. Symbiotic Augmentation (SA) which employs semantic segmentation yields a $\sim$2\% performance gain over our Inception-V3 \cite{szegedy2016rethinking} baseline surpassing \cite{Sarafianos_2018_ECCV}, the current state-of-the-art. For detailed performance comparison between varieties of ResNet \cite{he2016deep} and DenseNet \cite{huang2017densely} architectures on WIDER Attribute \cite{li2016human} dataset, readers are encouraged to refer to \cite{Sarafianos_2018_ECCV}.

Table \ref{tab:resultsBAPD} compares our proposed method with the state-of-the-art on Berkeley Attributes of People \cite{bourdev2011describing} dataset. Note that \cite{li2016human} leverages the context in the image while our method solely operates on the bounding box of each person, yet it still outperforms \cite{li2016human} with 2.6\% margin. Similar to WIDER Attribute \cite{li2016human} dataset, here utilizing semantic segmentation through our proposed Symbiotic Augmentation (SA) results in 2\% gain in AP over our already very competitive Inception-V3 \cite{szegedy2016rethinking} baseline. Detailed per-attribute results of our models are shown in Table \ref{tab:per_attr_fullbody}.

\begin{table*}
	\centering
	\begin{tabular}{lccccccccccc}
	\toprule
	Method & \rotatebox{90}{SSP+SSG} & \rotatebox{90}{SSP+SSG$^\star$} & \rotatebox{90}{\shortstack{Inception\\-V3: baseline}} & \rotatebox{90}{\shortstack{Symbiotic\\Aug. (SA)}} & \rotatebox{90}{SSP+SSG} & & \rotatebox{90}{SSP+SSG} & \rotatebox{90}{SSP+SSG$^\star$} & \rotatebox{90}{\shortstack{Inception\\-V3: baseline}} & \rotatebox{90}{\shortstack{Symbiotic\\Aug. (SA)}} & \rotatebox{90}{SSP+SSG}\\
	\midrule
	Dataset & \rotatebox{90}{CelebA} & \rotatebox{90}{CelebA} & \rotatebox{90}{CelebA} & \rotatebox{90}{CelebA} & \rotatebox{90}{LFWA} & & \rotatebox{90}{CelebA} & \rotatebox{90}{CelebA} & \rotatebox{90}{CelebA} & \rotatebox{90}{CelebA} & \rotatebox{90}{LFWA}\\
	\midrule
	 & \multicolumn{5}{c}{\textbf{Classification Accuracy(\%)}} & & \multicolumn{5}{c}{\textbf{Average Precision(\%)}}\\
	 \cmidrule{2-6}
	 \cmidrule{8-12}
    5 o Clock Shadow & 94.50 & 95.07 & 94.34 & 94.62 & 79.72 &  & 80.36 & 83.96 & 80.42 & 81.63 & 83.61\\
    Arched Eyebrows & 83.06 & 84.56 & 83.88 & 84.12 & 83.74 &  & 77.98 & 81.17 & 78.93 & 79.64 & 73.07\\
    Attractive & 82.25 & 83.28 & 82.21 & 82.27 & 80.89 &  & 91.14 & 92.50 & 91.18 & 91.36 & 83.83\\
    Bags Under Eyes & 85.42 & 86.15 & 85.26 & 85.60 & 85.09 &  & 67.68 & 70.05 & 67.24 & 67.96 & 95.19\\
    Bald & 98.79 & 99.02 & 98.92 & 98.95 & 92.76 &  & 76.43 & 84.03 & 79.11 & 79.40 & 71.09\\
    Bangs & 95.51 & 96.23 & 95.72 & 95.86 & 91.82 &  & 93.86 & 95.54 & 94.16 & 94.65 & 82.46\\
    Big Lips & 71.67 & 72.45 & 71.35 & 72.16 & 80.20 &  & 62.85 & 62.97 & 62.30 & 63.01 & 81.83\\
    Big Nose & 84.50 & 85.38 & 84.77 & 85.01 & 84.67 &  & 68.62 & 72.25 & 69.13 & 71.43 & 95.92\\
    Black Hair & 90.06 & 90.63 & 89.96 & 90.15 & 92.81 &  & 89.75 & 90.79 & 89.55 & 90.13 & 77.13\\
    Blond Hair & 95.82 & 96.30 & 95.90 & 95.94 & 97.72 &  & 91.45 & 92.73 & 91.54 & 91.67 & 78.77\\
    Blurry & 95.67 & 96.44 & 95.65 & 95.85 & 87.49 &  & 53.61 & 65.87 & 53.95 & 57.03 & 63.88\\
    Brown Hair & 89.25 & 89.95 & 88.42 & 88.46 & 82.72 &  & 76.58 & 78.97 & 75.22 & 75.18 & 83.76\\
    Bushy Eyebrows & 92.36 & 93.20 & 92.34 & 92.50 & 85.77 &  & 76.47 & 81.00 & 76.36 & 76.91 & 94.45\\
    Chubby & 95.61 & 96.02 & 95.80 & 95.94 & 77.66 &  & 56.24 & 62.54 & 59.63 & 62.39 & 76.48\\
    Double Chin & 96.28 & 96.61 & 96.23 & 96.47 & 81.86 &  & 58.42 & 63.92 & 58.49 & 61.86 & 85.80\\
    Eyeglasses & 99.27 & 99.67 & 99.51 & 99.48 & 92.79 &  & 98.43 & 99.20 & 98.52 & 98.49 & 86.96\\
    Goatee & 97.28 & 97.58 & 97.41 & 97.55 & 84.08 &  & 74.89 & 81.64 & 79.08 & 80.86 & 75.74\\
    Gray Hair & 98.22 & 98.37 & 98.16 & 98.30 & 89.24 &  & 77.32 & 80.49 & 77.65 & 79.32 & 71.69\\
    Heavy Makeup & 90.83 & 92.17 & 91.03 & 90.99 & 95.90 &  & 96.26 & 97.31 & 96.29 & 96.30 & 88.80\\
    High Cheekbones & 87.13 & 88.13 & 87.09 & 87.48 & 89.48 &  & 94.94 & 95.78 & 94.92 & 95.23 & 91.68\\
    Male & 97.67 & 98.51 & 98.00 & 98.08 & 94.42 &  & 99.59 & 99.83 & 99.69 & 99.73 & 99.08\\
    Mouth Slightly Open & 92.25 & 94.19 & 92.61 & 92.79 & 84.29 &  & 97.97 & 98.87 & 98.10 & 98.29 & 88.36\\
    Mustache & 96.96 & 97.01 & 96.94 & 97.16 & 94.01 &  & 64.14 & 67.94 & 65.45 & 67.01 & 86.11\\
    Narrow Eyes & 86.68 & 87.92 & 86.86 & 87.17 & 84.68 &  & 52.35 & 59.31 & 53.22 & 55.11 & 95.22\\
    No Beard & 95.66 & 96.52 & 95.77 & 95.74 & 83.63 &  & 99.74 & 99.82 & 99.76 & 99.79 & 94.98\\
    Oval Face & 77.83 & 76.83 & 77.15 & 77.50 & 77.89 &  & 66.25 & 63.84 & 65.40 & 65.75 & 87.21\\
    Pale Skin & 97.08 & 97.29 & 96.78 & 96.69 & 91.15 &  & 67.25 & 70.65 & 60.60 & 60.32 & 97.77\\
    Pointy Nose & 76.50 & 77.86 & 77.14 & 77.45 & 84.99 &  & 60.67 & 65.93 & 62.74 & 63.67 & 95.69\\
    Receding Hairline & 93.31 & 94.14 & 93.42 & 93.81 & 86.60 &  & 60.24 & 67.80 & 62.05 & 63.79 & 95.57\\
    Rosy Cheeks & 94.78 & 95.39 & 94.75 & 94.77 & 86.28 &  & 67.66 & 72.40 & 64.33 & 65.41 & 74.02\\
    Sideburns & 97.70 & 98.00 & 97.75 & 97.82 & 83.21 &  & 82.92 & 86.78 & 83.16 & 85.17 & 81.54\\
    Smiling & 91.92 & 93.39 & 92.00 & 92.45 & 92.51 &  & 97.97 & 98.62 & 98.07 & 98.23 & 97.00\\
    Straight Hair & 83.59 & 84.46 & 85.16 & 85.21 & 81.58 &  & 63.56 & 66.22 & 68.82 & 69.21 & 83.26\\
    Wavy Hair & 84.79 & 84.62 & 86.13 & 85.93 & 81.22 &  & 88.46 & 88.73 & 90.15 & 90.27 & 87.69\\
    Wearing Earrings & 89.99 & 90.94 & 90.41 & 90.56 & 95.23 &  & 83.40 & 85.71 & 84.79 & 85.18 & 89.11\\
    Wearing Hat & 98.78 & 99.11 & 99.07 & 99.07 & 91.08 &  & 92.87 & 95.89 & 95.21 & 95.59 & 75.11\\
    Wearing Lipstick & 93.58 & 94.56 & 93.61 & 93.88 & 95.19 &  & 98.67 & 99.10 & 98.70 & 98.76 & 90.52\\
    Wearing Necklace & 88.72 & 88.01 & 89.65 & 89.57 & 90.15 &  & 59.05 & 52.89 & 62.92 & 62.71 & 82.38\\
    Wearing Necktie & 97.15 & 97.02 & 97.17 & 97.12 & 83.87 &  & 86.81 & 87.51 & 87.45 & 88.31 & 94.47\\
    Young & 87.85 & 89.01 & 88.52 & 88.37 & 86.95 &  & 96.89 & 97.60 & 97.13 & 97.19 & 74.02\\
    \midrule
    \textbf{Avg.} & 91.16 & 91.80 & 91.32 & 91.47 & 87.13 &  & 78.74 & 81.45 & 79.28 & 80.10 & 85.28\\
	\bottomrule\\
	\end{tabular}
	\caption{Detailed per-attribute classification accuracy(\%) and average precision(\%) results of our proposed models for facial attribute prediction. Note that SSP+SSG$^\star$ indicates the experiment using pre-cropped images of CelebA.}
	\label{tab:per_attr_face}
\end{table*}

\begin{table}
	\centering
	\begin{tabular}{lcc}
	\toprule
    \multicolumn{3}{c}{\textbf{WIDER Attribute}\cite{li2016human}}\\
	\midrule	
	 & \shortstack{Inception-V3:\\baseline} & \shortstack{Symbiotic\\Augmentation (SA)}\\
	\cmidrule{2-3}
    Male & 95.60 & 96.64\\
    Long Hair & 86.98 & 89.25\\
    Sunglasses & 70.56 & 78.31\\
    Hat & 92.87 & 95.04\\
    T-shirt & 83.36 & 84.77\\
    Long Sleeve & 96.71 & 97.64\\
    Formal & 83.82 & 85.38\\
    Shorts & 91.96 & 93.87\\
    Jeans & 79.60 & 81.76\\
    Long Pants & 97.18 & 97.74\\
    Skirt & 85.74 & 87.65\\
    Face Mask & 76.51 & 79.18\\
    Logo & 91.07 & 90.87\\
    Stripe & 70.15 & 68.04\\
    \midrule
    \textbf{Avg.} & 85.86 & 87.58\\
    \midrule
    \multicolumn{3}{c}{\textbf{Berkeley Attributes of People} \cite{bourdev2011describing}}\\
	\midrule	
	 & \shortstack{Inception-V3:\\baseline} & \shortstack{Symbiotic\\Augmentation (SA)}\\
	\cmidrule{2-3}
    Is Male & 96.29 & 96.73\\
    Has Long Hair & 93.71 & 94.41\\
    Has Glasses & 79.57 & 88.41\\
    Has Hat & 92.97 & 96.31\\
    Has T-shirt & 86.28 & 88.15\\
    Has Long sleeves & 96.96 & 98.01\\
    Has Shorts & 95.43 & 95.82\\
    Has Jeans & 95.34 & 95.80\\
    Has Long Pants & 99.33 & 99.55\\
    \midrule
    \textbf{Avg.} & 92.87 & 94.80\\	
	\bottomrule\\
	\end{tabular}
	\caption{Detailed per-attribute AP(\%) results of our proposed models for person attribute prediction.}
	\label{tab:per_attr_fullbody}
\end{table}

\begin{figure*}
	\centering
	\includegraphics[width=1\textwidth]{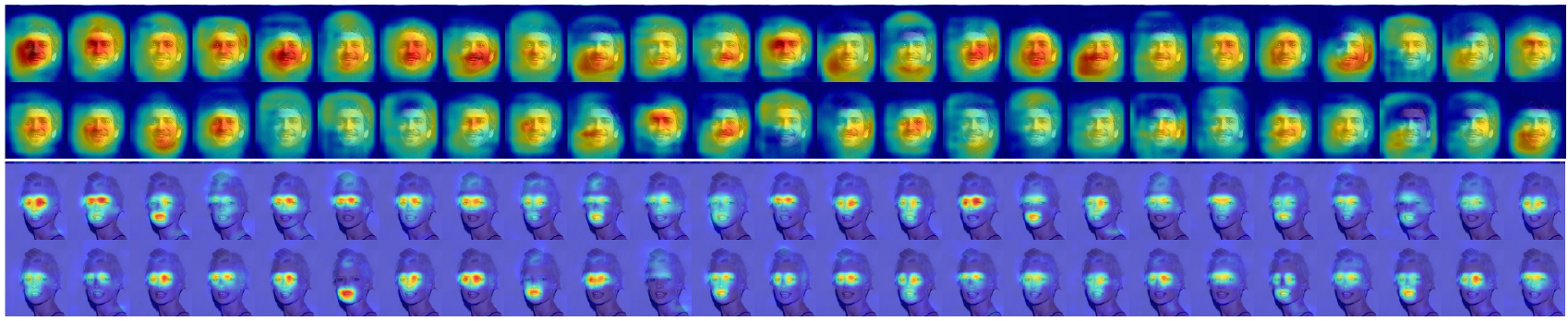}
	\caption{Top fifty activation maps of the last convolution layer sorted in descending order w.r.t the average activation values. Top: Basic attribute prediction model using global pooling. Bottom: SSP.}\label{fig:outputactivation}
\end{figure*}

\begin{figure*}
	\centering
	\includegraphics[width=0.95\textwidth]{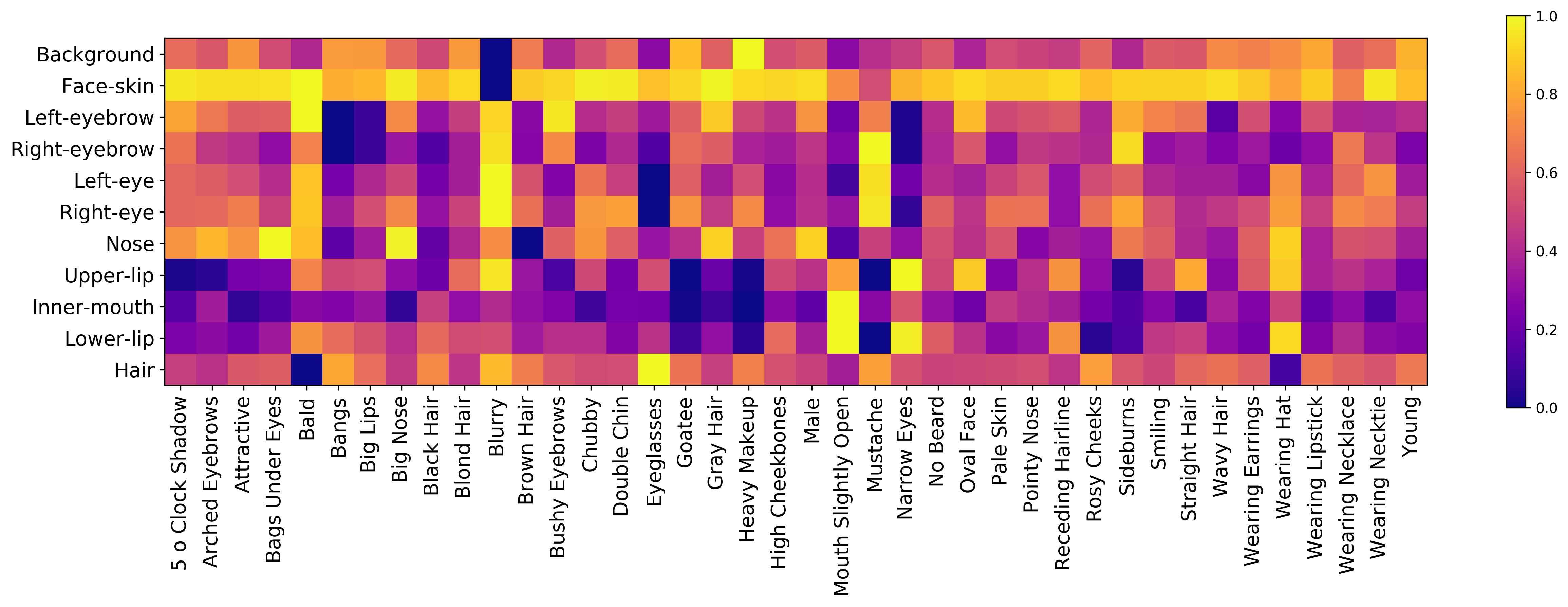}
\caption{Learned weights of $\Phi_{A}$ in Symbiotic Augmentation (SA), trained on CelebA and Helen. Note: 9 values associated with $3\times3$ kernels are averaged. For better visualization, values in each row are normalized between 0 and 1.}	
	\label{fig:weights_face_A}	
\end{figure*}

\begin{figure}
\begin{subfigure}{.49\textwidth}
  \centering
  \includegraphics[width=.99\textwidth]{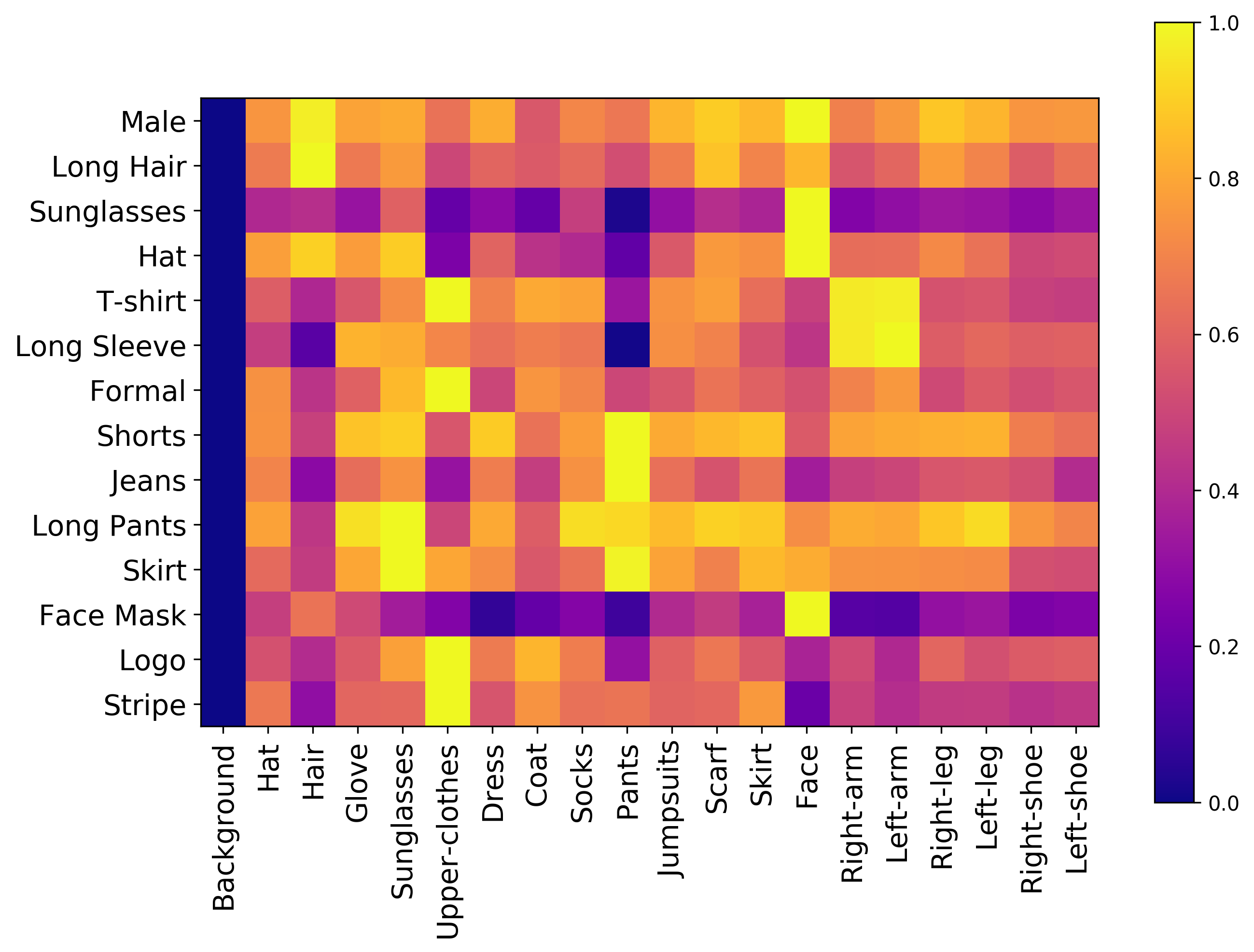}
  \caption{$\Phi_{S}$}
  \label{fig:weights_body_S}
\end{subfigure}\\
\bigskip
\begin{subfigure}{.49\textwidth}
  \centering
  \includegraphics[width=.99\textwidth]{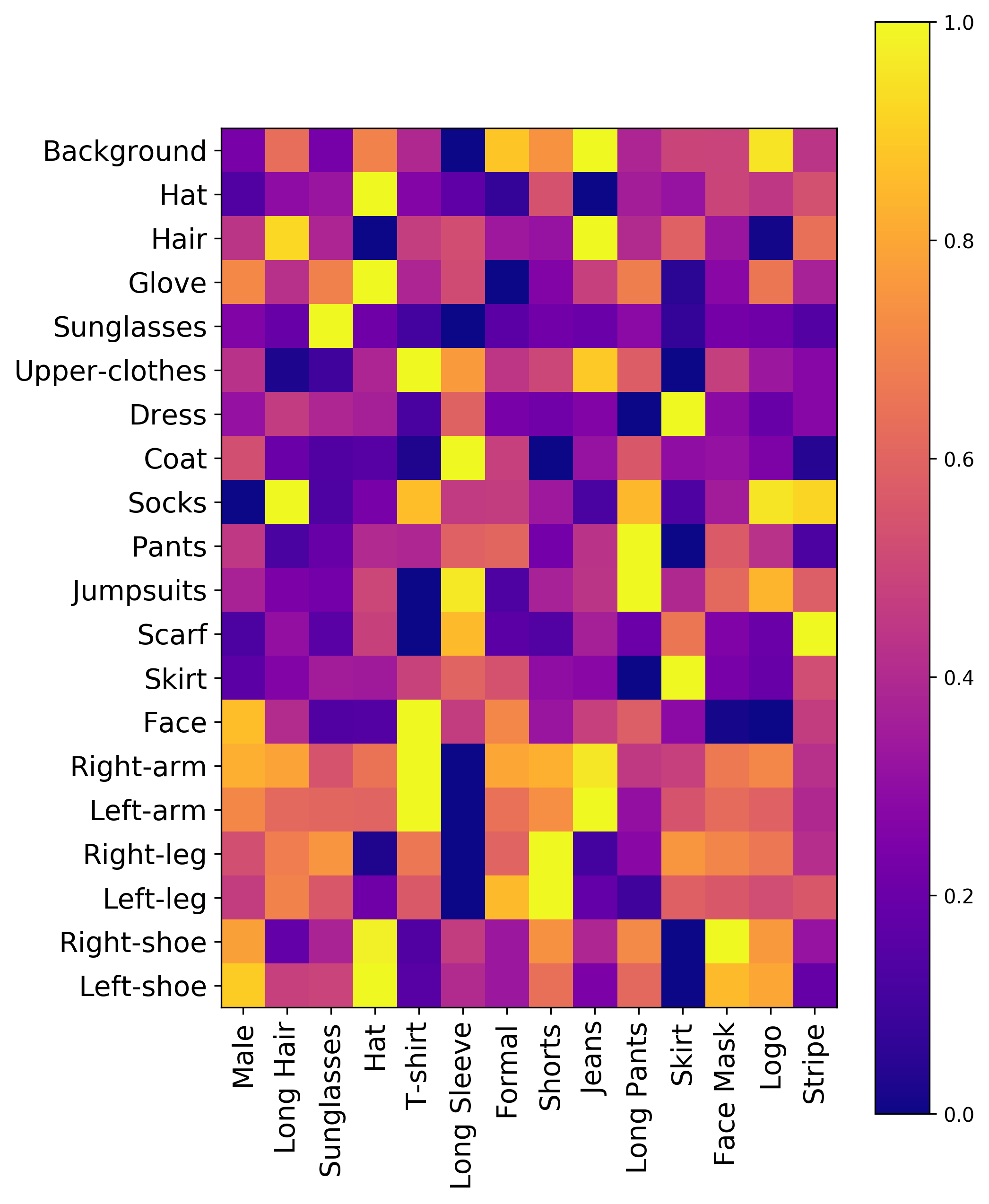}
  \caption{$\Phi_{A}$}
  \label{fig:weights_body_A}
\end{subfigure}
\caption{Learned weights of embedding convolution layers in Symbiotic Augmentation (SA), trained on WIDER and LIP. Note: 9 values associated with $3\times3$ kernels are averaged. For better visualization, values in each row are normalized between 0 and 1.}
\end{figure}

\begin{figure}
	\centering
	\includegraphics[width=0.45\textwidth]{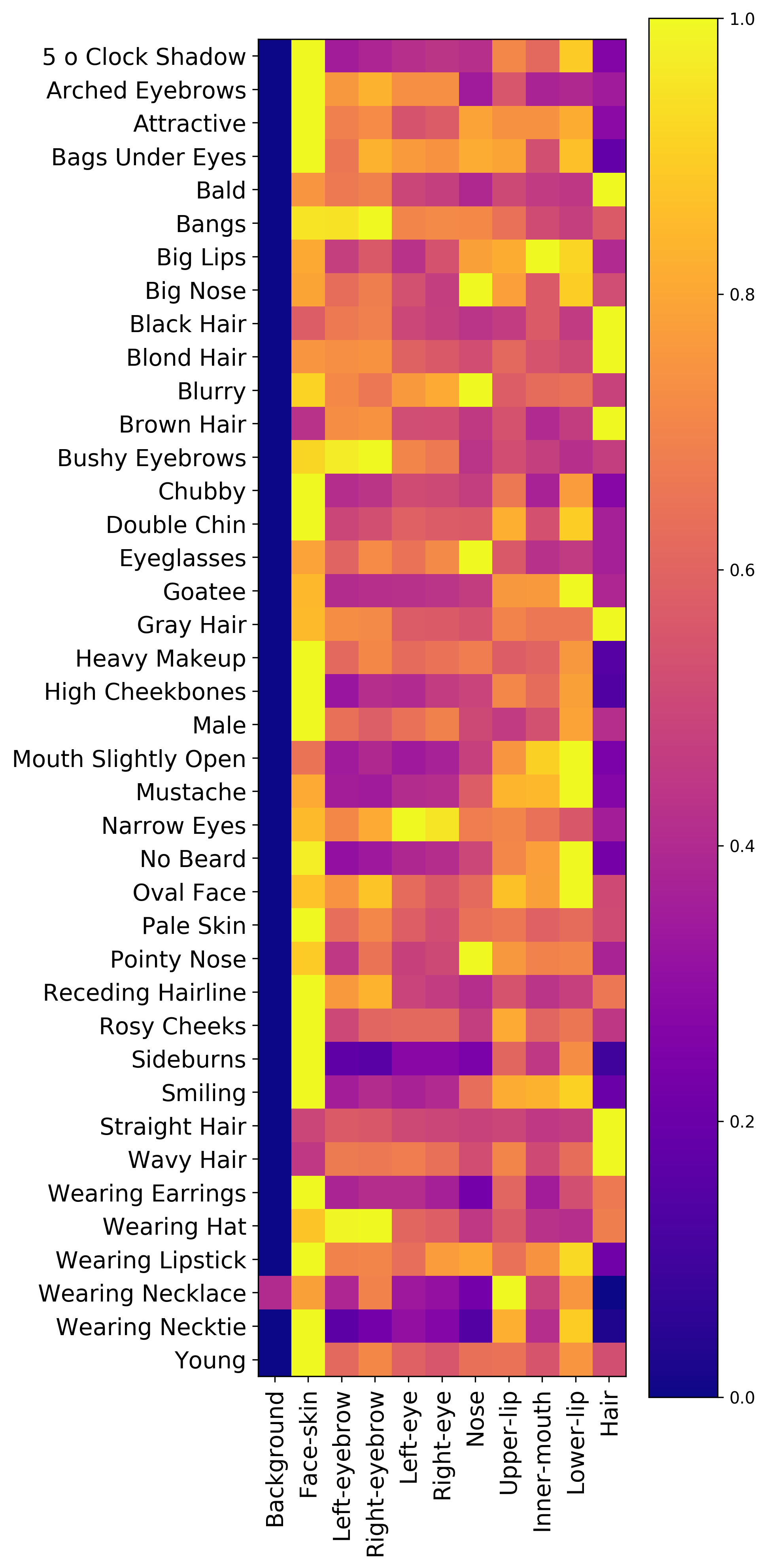}
\caption{Learned weights of $\Phi_{S}$ in Symbiotic Augmentation (SA), trained on CelebA and Helen. Note: 9 values associated with $3\times3$ kernels are averaged. For better visualization, values in each row are normalized between 0 and 1.}	
	\label{fig:weights_face_S}
\end{figure}

\subsection{Visualizations}
Unlike the global average pooling which equally affects a rather large spatial domain, we expect SSP to generate activations that are semantically aligned. To evaluate our hypothesis, in Figure \ref{fig:outputactivation}, we show the activations for the top fifty channels of the last convolution layer. Top row corresponds to our basic network with global average pooling, while the bottom row is generated when we replace global average pooling with SSP. We observe that, activations generated by SSP are clearly more localized than those obtained from the global average pooling.

To better understand how attribute prediction and semantic segmentation models have learned their corresponding tasks, we visualize the embedding convolution layers $\Phi_{S}$ and $\Phi_A$ (ref. Figure \ref{fig:improved_ssg}) for simultaneously training of CelebA \cite{liu2015faceattributes} (original image set) with Helen face \cite{le2012interactive}, and WIDER Attribute \cite{li2016human} with LIP \cite{gong2017look}. Figure \ref{fig:weights_face_S} shows how for each facial attribute (vertical axis), network has learned to employ different semantic regions of face (horizontal axis) in order to predict attributes. Note that these weights are learned through back-propagation and are not hard coded, yet they reveal very interesting observations. First, almost all the attributes give ``background'' the lowest importance, except attribute ``Wearing Necklace'' which makes sense as neck falls outside the face region and counted as background in Helen face dataset \cite{le2012interactive}. Second, the learned importance for the majority of attributes are aligned with human expectations. For instance, all the hair-related attributes are inferred with the most attention of the model being paid to the ``Hair'' region. The same is true for ``Big Nose'', ``Pointy Nose'' and ``Eyeglasses'' as the model learns to focus on the ``Nose'' region. Figure \ref{fig:weights_face_A} illustrates $\Phi_{A}$ for the reverse problem where attributes are supposed to improve semantic face parsing. Figure \ref{fig:weights_body_S} and \ref{fig:weights_body_A} show the learned weights of the embedding convolution layer for person attribute prediction and human semantic parsing tasks. 

We observe that simultaneously training for attribute prediction and semantic segmentation within Symbiotic Augmentation framework, in addition to the performance gains, provides us with meaningful tools to study how a complex deep neural network infers and relates different semantic labels across multiple tasks.

\subsection{Attribute Prediction for Semantic Segmentation}\label{sec:attributes_for_segmentations}
In this work, we have established how semantic segmentation can be used to improve person-related attribute prediction. What if we reverse the roles. Can attributes improve semantic parsing problem? To evaluate this, we focus on facial attributes and compare the performance of semantic face parsing on Helen face \cite{le2012interactive}. We consider three scenarios. First, initializing Inception-V3 \cite{szegedy2016rethinking} backbone with ImageNet \cite{russakovsky2015imagenet} pre-trained weights. Second, training a baseline attribute prediction network on CelebA \cite{liu2015faceattributes} and using the corresponding weights, once training finished, to initialize semantic face parsing network. Third, training facial attribute and semantic face parsing simultaneously through Symbiotic Augmentation (SA) framework. For the sake of simplicity, solely in this experiment, SA only uses the final activations of the CNN backbone instead of concatenating them with intermediate feature maps as shown in Figure \ref{fig:weightsharing}. We observed that upgrading to full SA model boosts mean class accuracy by $\sim$5\% and also achieves similar mean IoU. Table \ref{tab:results_on_attributes_for_segmentations1} shows that pre-training on image-level facial attribute annotations delivers a large performance gain over ImageNet based initialization. This shows that there exists an \textit{interrelatedness} between attribute prediction and semantic segmentation. Furthermore, it suggests that while collecting annotations for semantic parsing is laborious and expensive, instead one can use relevant image-level attribute annotations to initialize a semantic parsing model. The last row in each block of the Table \ref{tab:results_on_attributes_for_segmentations1} demonstrates how training facial attributes and semantic face parsing jointly, through our proposed Symbiotic Augmentation (SA), can further push the performance boundary with significant margin. Therefore, it is easy to see that when few training instances are available, indeed image-level facial attribute labels can serve as an effective source of weak supervision to improve semantic face parsing task. In fact such \textit{interrelatedness} plays a major role in allowing us to successfully unify semantic segmentation and attribute predictions networks (ref. Section \ref{sec:methodology}) without sacrificing the performance. Jointly training on LIP \cite{gong2017look} and WIDER Attribute \cite{li2016human}, we did not observe meaningful gain in semantic segmentation task on LIP \cite{gong2017look}. We hypothesize that, this is due to the fact that LIP \cite{gong2017look} itself already has huge ($\sim$30,000 instances) number of training annotations. In order to confirm this, conducting an experiment where only a small portion of LIP \cite{gong2017look} training instances are used is needed.

\begin{table*}[htbp]
    \small
	\centering
	\begin{tabular}{lccccccccccc|c}
	\toprule
	\multicolumn{12}{c}{\textbf{Intersection over Union(\%)}}\\
	\midrule	
	Method & bkg & face skin & l-eyebrow & r-eyebrow & l-eye & r-eye & nose & u-lip & i-mouth & l-lip & hair & \textbf{Avg.}\\
	\midrule
    init: ImageNet & 92.97 & 85.58 & 46.90 & 48.33 & 55.39 & 55.91 & 84.24 & 43.77 & 59.21 & 55.19 & 71.99 & 63.58\\
    init: CelebA & 93.20 & 86.40 & 51.31 & 51.11 & 56.22 & 58.81 & 84.82 & 49.32 & 60.01 & 58.95 & 73.13 & 65.75\\
    SA & \textbf{94.25} & \textbf{88.24} & \textbf{59.29} & \textbf{58.11} & \textbf{62.45} & \textbf{67.22} & \textbf{87.96} & \textbf{51.05} & \textbf{69.66} & \textbf{70.32} & \textbf{75.77} & \textbf{71.29}\\
	\toprule
	\multicolumn{12}{c}{\textbf{Class Accuracy(\%)}}\\
	\midrule	
	method & bkg & face skin & l-eyebrow & r-eyebrow & l-eye & r-eye & nose & u-lip & i-mouth & l-lip & hair & \textbf{Avg.}\\
	\midrule
    init: ImageNet & 96.04 & 94.21 & 56.02 & 60.95 & 67.61 & 67.62 & 90.69 & 58.25 & 74.73 & 66.12 & 83.36 & 74.14\\
    init: CelebA & 95.96 & 94.09 & 63.31 & 67.71 & 67.30 & 69.79 & 90.06 & 66.80 & 75.27 & 72.83 & 85.22 & 77.12\\
    SA & \textbf{97.02} & \textbf{95.47} & \textbf{69.89} & \textbf{74.97} & \textbf{72.12} & \textbf{77.21} & \textbf{92.43} & \textbf{66.96} & \textbf{76.88} & \textbf{81.60} & \textbf{84.67} & \textbf{81.07}\\
	\bottomrule\\
	\end{tabular}
	\caption{Effect of leveraging image-level attribute supervision for semantic face parsing, evaluated on the test split of Helen face \cite{le2012interactive}\cite{smith2013exemplar}. Here, all the models were trained with the input image resolution of $448\times448$.}
	\label{tab:results_on_attributes_for_segmentations1}
\end{table*}

\section{Conclusion}\label{sec:conclusion}
Aligned with the trend of part-based attribute prediction methods, we proposed employing semantic segmentation to improve person-related attribute prediction. Specifically, we jointly learn attribute prediction and semantic segmentation in order to mainly transfer localization cues from the latter task to the former. To guide the attention of our attribute prediction model to the regions which different attributes naturally show up, we introduced SSP and SSG. While SSP is used to restrict the aggregation procedure of final activations to regions that are semantically consistent, SSG carries the same notion but applies it to the earlier layers. We then demonstrated that there exists a single unified architecture that can mimic the behavior of SSP and SSG, depending on where in the network architecture it is being used. We evaluated our proposed methods on CelebA, LFWA, WIDER Attribute and Berkeley Attributes of People datasets and achieved state-of-the-art performance. We also showed that attributes can improve semantic segmentation (in case of few training instances) when properly used through our Symbiotic Augmentation (SA) framework. We hope to encourage future research works to invest more in the interrelatedness of these two problems.

\ifCLASSOPTIONcompsoc
  \section*{Acknowledgments}
  This material is based upon work supported by the National Science Foundation under Grant No. 174143 and the Office of the Director of National Intelligence (ODNI), Intelligence Advanced Research Projects Activity (IARPA), via IARPA R\&D Contract No. D17PC00345. The views and conclusions contained herein are those of the authors and should not be interpreted as necessarily representing the official policies or endorsements, either expressed or implied, of the ODNI, IARPA, or the U.S. Government. The U.S. Government is authorized to reproduce and distribute reprints for Governmental purposes notwithstanding any copyright annotation thereon.  
\else
  \section*{Acknowledgment}
\fi

\ifCLASSOPTIONcaptionsoff
  \newpage
\fi

\bibliographystyle{IEEEtran}
\bibliography{references}

\begin{IEEEbiography}[{\includegraphics[width=1in,height=1.25in,clip,keepaspectratio]{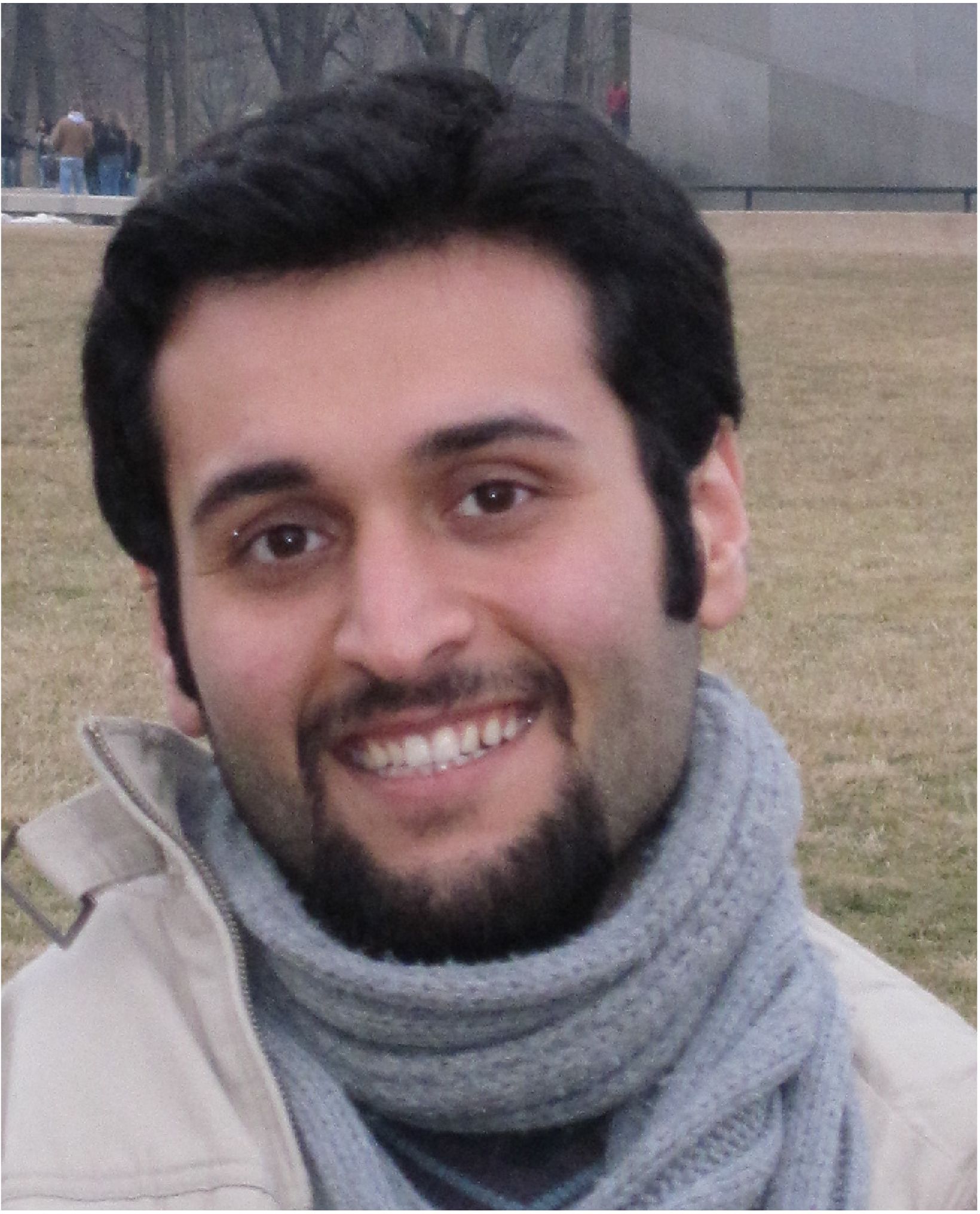}}]{Mahdi M. Kalayeh}
received his B.Sc. from Tehran Polytechnic (Amirkabir University of Technology) in 2009 and M.Sc. from Illinois Institute of Technology (IIT) in 2010, both in Electrical Engineering. In 2019, Mahdi graduated with Ph.D. in Computer Science from Center for Research in Computer Vision (CRCV) at the University of Central Florida. His research is on the intersection of Computer Vision and Machine Learning, specifically, it includes Deep Learning, Visual Attribute Prediction, Semantic Segmentation, Complex Event and Action Recognition, Object Recognition and Scene Understanding. Mahdi has published several papers in conferences and journals such as CVPR, ACMMM, and PAMI. He has also served as a reviewer for peer-reviewed conferences and journals including CVPR, ICCV, ECCV, ACCV, IJCV, IEEE Transactions on Image Processing, and IEEE Transactions on Multimedia. Mahdi is currently a Senior Research Scientist at Netflix.
\end{IEEEbiography}
\vfill
\begin{IEEEbiography}[{\includegraphics[width=1in,height=1.25in,clip,keepaspectratio]{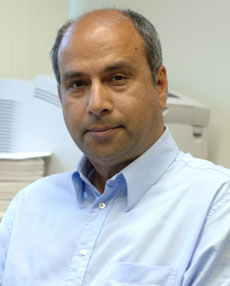}}]{Mubarak Shah}
Mubarak Shah, the Trustee chair professor of computer science, is the founding director of the Center for Research in Computer Vision at the University of Central Florida (UCF). He is an editor of an international book series on video computing, was editor-in-chief of Machine Vision and Applications journal, and an associate editor of ACM Computing Surveys journal. He was the program cochair of CVPR 2008, an associate editor of the IEEE T-PAMI, and a guest editor of the special issue of the International Journal of Computer Vision on Video Computing. His research interests include video surveillance, visual tracking, human activity recognition, visual analysis of crowded scenes, video registration, UAV video analysis, and so on. He is an ACM distinguished speaker. He was an IEEE distinguished visitor speaker for 1997-2000 and received the IEEE Outstanding Engineering Educator Award in 1997. In 2006, he was awarded a Pegasus Professor Award, the highest award at UCF. He received the Harris Corporations Engineering Achievement Award in 1999, TOKTEN awards from UNDP in 1995, 1997, and 2000, Teaching Incentive Program Award in 1995 and 2003, Research Incentive Award in 2003 and 2009, Millionaires Club Awards in 2005 and 2006, University Distinguished Researcher Award in 2007, Honorable mention for the ICCV 2005 Where Am I? Challenge Problem, and was nominated for the Best Paper Award at the ACM Multimedia Conference in 2005. He is a fellow of the IEEE, AAAS, IAPR, and SPIE.
\end{IEEEbiography}

\end{document}